\pdfoutput=1

\documentclass[11pt]{article}

\usepackage[final]{acl}

\usepackage{times}
\usepackage{latexsym}

\usepackage[T1]{fontenc}

\usepackage[utf8]{inputenc}

\usepackage{microtype}

\usepackage{inconsolata}

\usepackage{graphicx}

\usepackage{amsmath}
\usepackage{pifont}
\newcommand{\xmark}{\ding{55}}

\usepackage{amssymb}
\usepackage{xcolor}
\usepackage{color}
\usepackage{colortbl}
\usepackage{url}
\usepackage{bm}
\usepackage{multirow}
\usepackage{booktabs}
\usepackage{CJKutf8}
\newcommand{\ja}[1]{\begin{CJK}{UTF8}{ipxm}#1\end{CJK}}

\definecolor{lblue}{RGB}{178,216,255}
\definecolor{llblue}{RGB}{224,239,255}
\definecolor{pink}{RGB}{255,190,230}
\definecolor{ppink}{RGB}{255,224,255}

\title{Comprehensive Evaluation on Lexical Normalization:\\
Boundary-Aware Approaches for Unsegmented Languages}

\author{
Shohei Higashiyama \and Masao Utiyama\\
National Institute of Information and Communications Technology, Kyoto, Japan\\
\texttt{\{shohei.higashiyama,mutiyama\}}\texttt{@nict.go.jp}\\
}

\begin{document}
\maketitle
\begin{abstract}
Lexical normalization research has sought to tackle the challenge of processing informal expressions in user-generated text, yet the absence of comprehensive evaluations leaves it unclear which methods excel across multiple perspectives.
Focusing on unsegmented languages, we make three key contributions: (1) creating a large-scale, multi-domain Japanese normalization dataset, (2) developing normalization methods based on state-of-the-art pre-trained models, and (3) conducting experiments across multiple evaluation perspectives. Our experiments show that both encoder-only and decoder-only approaches achieve promising results in both accuracy and efficiency.\footnote{Our dataset and code will be available at \url{https://github.com/shigashiyama/jmln}.}
\end{abstract}
\thispagestyle{plain}

\section{Introduction}
User-generated text (UGT) is invaluable textual content produced by users' activities on web platforms such as review sites and social media.
UGT's informality---its frequent use of colloquial expressions---poses substantial challenges to accurate analysis in natural language processing applications.
Consequently, researchers have explored lexical normalization (LN),
the task of converting non-standard word forms into standard ones.
LN has been actively studied particularly for space-delimited languages such as European languages~\cite{baldwin2015,van-der-goot-etal-2021-multilexnorm}.
In this study, we investigate LN for \textit{unsegmented languages in writing}, focusing primarily on Japanese.

Major issues in existing LN research are summarized as a lack of comprehensive evaluation, leaving unclear which methods excel under 
 different evaluation aspects. 
Specifically, (i) comparative evaluations of recent model architectures are absent, and (ii) multi-perspective analyses---examining required training data size, inference cost, and domain-specific accuracy across diverse domains---have not been conducted.
These issues are common to LN research but are particularly severe for underexplored unsegmented languages.
This study addresses these gaps through three key contributions: (1) dataset construction, (2) method development based on cutting-edge pre-trained models, and (3) comprehensive experiments.

First, we introduce the Japanese Multi-Domain Lexical Normalization Dataset (JMLN), a large collection of 21,402 sentences drawn from a variety of UGT sources.
JMLN's size exceeds existing Japanese LN datasets~\cite{higashiyama-etal-2021-user,kondo-etal-2025-text}, and its domain diversity surpasses that of any current LN datasets.
These properties enable both the development of methods suitable for Japanese and multi-perspective evaluation.

Second, we develop LN methods based on three modern Transformer~\cite{vaswani-etal-2017} architectures---encoder-only, encoder-decoder, and decoder-only---including a novel encoder-based infilling approach, as well as variants of generative approaches.
While these \textit{boundary}-\textit{aware} methods are tailored for unsegmented languages, they remain broadly applicable.

Third, we evaluate these methods on JMLN and an existing Thai dataset.
Multi-perspective experiments on JMLN yield in-depth insights into methods’ characteristics and trade-offs, while experiments on the Thai dataset further validate cross-lingual applicability and generalizability.

Our evaluation reveals three main findings. First, compact encoder-only models deliver the highest inference throughput, and decoder-only models excel in normalization recall---while both yield high normalization precision. Second, training models on 4k--8k sentences yields reasonable precision of around 0.7, and cutting-edge decoder-only models deliver superior recall even with fewer instances. Third, domains rich with unknown informal words exhibit low performance, especially the typo-correction domain.

\section{Related Work}
Text normalization has been studied for some representative purposes: 
mapping dialectal variants to standard language \cite{kuparinen-etal-2023-dialect},
modernizing historical writings \cite{bollmann-2019-large},
and verbalizing semiotic expressions for text-to-speech (TTS) synthesis \cite{zhang-2019-neural}.
Specifically, lexical normalization (LN) refers to the normalization task of converting UGT at the lexical level.

To date, research on UGT normalization can be broadly divided into three categories based on the dominant methodologies of each period: rule-based and statistical methods \cite{aw-etal-2006-phrase,choudhury2007investigation,han2011}, pre-Transformer neural methods \cite{chrupala-2014-normalizing,ikeda2016,lusetti-etal-2018-encoder,lourentzou2019adapting}, and Transformer-based methods \cite{muller-etal-2019-enhancing,samuel-straka-2021-ufal,bucur-etal-2021-sequence,bikaun-etal-2024-maintnorm}.

To achieve robust word segmentation (WS) for UGT, previous studies have often addressed LN jointly with WS, as exemplified by research on Japanese \cite{sasano2013,kaji2014,saito2014a,saito2017a,higashiyama-etal-2021-text}, Chinese \cite{wang2013,qian2015}, and Thai \cite{haruechaiyasak-2013-lextoplus}.
Some recent studies have adopted Transformer masked language models (MLMs) 
with a text-editing approach \cite{ueda-etal-2023-kwja} and a two-step approach---first detecting informal tokens and then predicting formal tokens \cite{pankam2023two}.

Decoder-only Transformer models have recently made remarkable advances and have been applied to text normalization for TTS \cite{zhang2024chatboringproblemsstudying,shen2024tnformer} and other sequence transduction tasks \cite{kaneko-okazaki-2023-reducing,shi-etal-2024-lexical}.
However, these models remain underexplored in UGT normalization, resulting in a lack of comparative evaluation of state-of-the-art models for this task.

Regarding evaluation domains, many studies have focused on building short message and social media datasets for European \cite{choudhury2007investigation,han2011,baldwin2015,plank-etal-2020-dan,van-der-goot-etal-2021-multilexnorm} and Asian languages \cite{kaji2014,limkonchotiwat-etal-2021-handling,nguyen-etal-2024-vilexnorm,kondo-etal-2025-text}, which has led to extensive model development and evaluation in these domains.
Some studies have focused on other domains
such as blog and Q\&A site~\cite{higashiyama-etal-2021-user}, and maintenance short text~\cite{bikaun-etal-2024-maintnorm}.
However, cross-domain evaluation research covering three or more domains remains scarce.

\section{Japanese Dataset Construction} \label{sec:data_const}
For our primary target language, Japanese, existing datasets~\cite{kaji2014,osaki2017,higashiyama-etal-2021-user,kondo-etal-2025-text} have limited domain diversity and size---covering one or two domains with approximately 1,000 to 6,000 sentences/posts.
In this study, we have constructed JMLN, as a large-scale dataset sourced from a variety of UGT.

\paragraph{Data Sources and Size}
We sampled original texts from various sources: Q\&A site, blog site, review site, recipe site, video site, online forum, and social media platform, as well as Wikipedia edit history and conversation transcriptions (details in Appendix~\ref{sec:data_src}).
The constructed dataset includes 21,402 sentences with 8,885 normalization instances, i.e., non-standard and standard form pairs (details in Appendix \ref{sec:data_stat}).
The large data size and domain diversity of our dataset are advantages over existing Japanese datasets, enabling multi-perspective evaluations, as shown in \S\ref{sec:exp}.

\paragraph{Basic Designs}
We followed~\citet{higashiyama-etal-2021-user}'s annotation criteria. The annotation information includes word boundaries based on the short-unit word criterion~\cite{maekawa2014}, as well as word attributes such as part-of-speech, lemma, predefined word categories (details in Appendix~\ref{sec:cate}), and standard forms of non-standard words.\footnote{Non-standard forms are those with distinctive orthographic features whose frequency in the reference corpus falls below a threshold, whereas standard forms are those whose frequency exceeds a threshold. See details in Appendix \ref{sec:sfrom_def}.}
Thus, the dataset supports both LN and Japanese morphological analysis, which involves word boundary detection, part-of-speech tagging, and lemmatization. Nevertheless, this study focuses on the evaluation of LN, leaving morphological analysis tasks outside its scope.

\paragraph{Annotation Process}
As part of data preparation, we extracted sentences with a reasonable length (10--300 characters) as a candidate set from each of the original 14 datasets.
Four experienced annotators, one of whom was a manager, at a data annotation company then carried out the annotation process as follows.
\begin{description}
\setlength{\parskip}{0cm} 
\setlength{\itemsep}{0.1cm}
\item[$\bullet$]\!\!Sentence selection: 
From each candidate set for each original dataset, annotators intentionally selected sentences containing UGT-specific category words (Appendix \ref{sec:cate}). 

\item[$\bullet$]\!\!Word information annotation: Annotators annotated the selected sentences with word information by modifying auto-analyzed results by a morphological analyzer, MeCab~\cite{kudo-etal-2004-applying} with UniDic (\texttt{unidic-cwj-3.1.0})~\cite{den2009}.\footnote{\url{https://clrd.ninjal.ac.jp/unidic/}}

\item[$\bullet$]\!\!Standard form annotation: We adopted separated steps for standard form (SForm) annotation---(1) recognizing non-standard words by assigning variant-form type categories (Appendix \ref{sec:cate}), (2) assigning an SForm ID to each non-standard word, and (3) associating a set of valid standard forms with each SForm ID. While the first step was performed by annotators, the second and third steps were performed by the annotation manager.
\end{description}

\paragraph{Inter-Annotator Agreement}
During the annotation process, 
manager A selected a total of 240 sentences from the candidate sets, and then two annotators---B and either C or D---independently annotated those sentences with boundaries and attributes.
Inter-annotator agreement (IAA) for non-standard word recognition on these sentences, as measured by F${}_1$ score, was 0.836 (see Appendix~\ref{sec:iaa}).\footnote{\citet{plank-etal-2020-dan} and \citet{van-der-goot-etal-2020-norm} used Cohen's kappa to evaluate IAA in informal word classification, based on whether each given word is normalized or not. This metric is not directly applicable to unsegmented languages.}
The datasets' high annotation consistency and large size suggest its usefulness, and this is further demonstrated by the experiments in \S\ref{sec:exp}, where the evaluated models achieve high normalization accuracy.

\section{Task Definition} \label{sec:task_def}
Following previous studies~\cite{sasano2013,baldwin2015}, we define LN as a task of \textit{boundary-aware} span extraction and conversion, in which a system not only generates a normalized text but also identifies the original spans of each informal words (or phrases).
An example input-output pair for our task is shown in Figure~\ref{fig:encoder}.

This task can be carried out independently of the tokenization unit and applied to any LN dataset, as long as informal-to-formal alignments are annotated.\footnote{Otherwise, a non-trivial alignment processing is needed. We provide an alignment examples in Appendix \ref{sec:align}.}
In contrast to the text-to-text conversion task~\cite{ikeda2016}, the boundary-aware task allows fine-grained evaluation at the normalization-span level and provides better interpretability of system outputs.
This is because span-level evaluation metrics directly assess the validity of each normalization instance, leading to more robust evaluation across sentences with varying densities of non-standard tokens.
Further discussion on the task definition is provided in Appendix \ref{sec:task_discussion}


Formally, an LN system takes as input a source sentence, namely, a sequence of $n$ character (or subword) tokens $\bm{x}=\bm{x}_{0:n}=[x_0,\dots,x_{n-1}]$, and is required to predict the set of non-standard word spans and their standard forms $P=\{(b,e,s)\}$. Here, $(b, e)$ ($0\!\leq\!b\!\leq\!e\!\leq\!n$) indicates a span of an non-standard word $x_{b:e}$ with length $e-b$ in the source sentence, and $s$ indicates its standard form.\footnote{A gold-standard normalization instance $i$ can have multiple standard forms and is thus represented as $(b_i,e_i,\{s_{i,k}\}_k)$.}
Each standard form $s$ is a string with length $\geq 0$, \\
where length $=0$ indicates that the non-standard word should be deleted in the normalized sentence. When $b=e$, a zero-length span indicates some token(s) should be inserted into the position $b$.

\section{Methods} \label{sec:methods}
We present boundary-aware LN methods based on three Transformer architectures: encoder-only, encoder-decoder, and decoder-only.
Our encoder-based infilling approach is a novel method for unsegmented languages, and comparing multiple approaches across different architectures offers a valuable, novel evaluation.

\subsection{Infilling Approach}
\begin{figure}[t]
\centering
\includegraphics[width=1\linewidth]{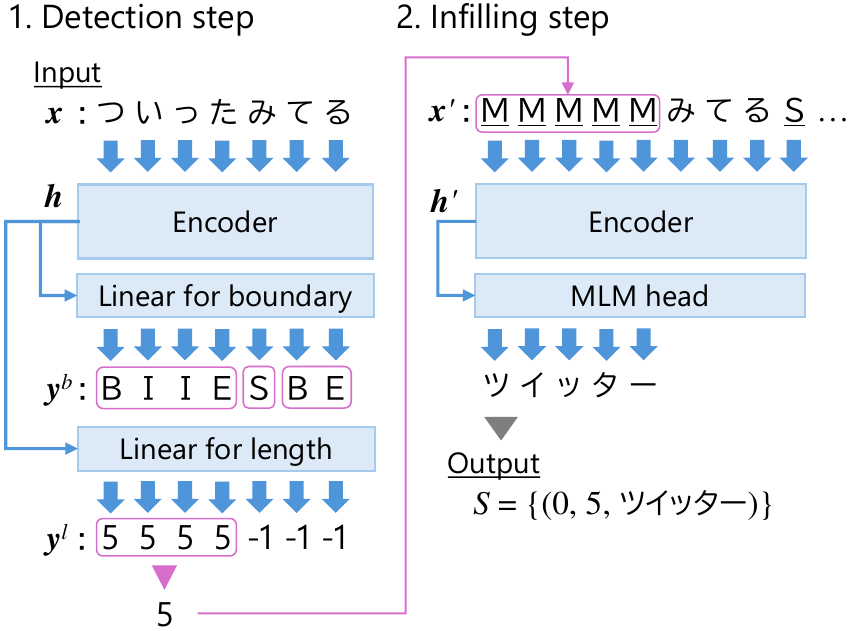}
\caption{Flow of our detect\&infill approach for an input text ``\ja{ついったみてる},'' which means ``(I'm) looking at Twitter.'' ``\underline{M}'' and ``\underline{S}'' represent the \texttt{MASK}  and \texttt{SEP} token, respectively. The original characters ``\ja{ついった}'' follow the \texttt{SEP} token, but are omitted in the Figure.}
\label{fig:encoder}
\end{figure}


Among encoder-based methods, including text editing~\cite{ueda-etal-2023-kwja} and MLM infilling~\cite{muller-etal-2019-enhancing}, the latter directly leverage the capabilities of pretrained MLMs to insert any token from the vocabulary.
As a representative study, \citet{muller-etal-2019-enhancing} proposed a two-step approach for space-delimited text, which predicts the infilling lengths and infilling tokens from subword-tokenized text.
While we follow the two-step \textit{detect-and-infill} framework, we propose a solution tailored to unsegmented languages. 
Our method jointly predicts word boundaries and infilling lengths from the input character sequence, thereby identifying non-standard word spans and their corresponding normalized spans, without the explicit alignment step \cite{muller-etal-2019-enhancing}.


As shown in Figure~\ref{fig:encoder}, the detailed workflow of our approach is as follows.
In the \textit{detection} step, the encoder takes as input a character token sequence $\bm{x}_{0:n}$ and output the sequence of token hidden representations $\bm{h}_{0:n}$. Then, a linear layer for boundary prediction and that for length prediction predict a chunk boundary tag sequence $\bm{y}^b_{0:n}$ and a length value sequence $\bm{y}^l_{0:n}$, respectively.
A series of boundary tags\footnote{We adopt the BIESO tagging schema in our experiments.} (e.g., $y^b_{0:4}=[\verb|B|, \verb|I|, \verb|I|, \verb|E|]$) identifies a chunk corresponding a standard or non-standard word.
A positive length value (e.g., ``5'') for the specified non-standard word chunk (e.g., $x_{0:4}$) indicates the numbers of tokens comprising standard forms that should be filled in the later step, the value ``0'' indicates that the chunk should be removed, and the value ``-1'' indicates that the chunk has no need for normalization.
Notably, we estimate a length value per character token, so an non-standard word chunk of $m$ characters yields $m$ redundant length values (e.g., $y^l_{0:4}=[5, 5, 5, 5]$). We determine a single length value by taking a majority vote within the chunk.
Thus, the combination of two sequences specifies non-standard word spans and the lengths of their standard forms.

In the \textit{infilling} step, the encoder takes as input the source text $\bm{x}^\prime$, in which tokens in non-standard word spans are replaced by the \texttt{MASK} tokens, and the MLM head predicts appropriate tokens for the masked positions from the hidden representations $\bm{h}^\prime$.
We apply input extension to the masked source text by concatenating the original characters of the specified non-standard tokens, with \texttt{SEP} tokens inserted between them, similarly to existing sequence transduction methods~\cite{qiang2021lsbert,he-etal-2023-umrspell}.

We refer to the above method as the \textsc{Full-Seg} approach.
During training, the model is optimized using the sum of cross-entropy losses over multiple subtasks from both steps.

As a variant of this approach, the detection step can be performed eigher with full-word segmentation or, alternatively, with partial word segmentation, which requires only the boundaries of (continuous) non-standard words in a sentence. The latter approach, referred to as \textsc{Part-Seg}, is introduced in Appendix~\ref{sec:enc_variants}.
We report a comparison among variants, including these two, in Appendix~\ref{sec:anal_enc}.

\subsection{Generative Approaches}
Encoder-decoder models have been extensively used in normalization research for text-to-text conversion, which we refer to the \textit{plain full-text} (\textsc{Plain}) approach.
To eliminate the informal-to-formal alignment step required for this approach, \citet{bikaun-etal-2024-maintnorm} generates outputs in which non-standard words and their corresponding normalized forms are each surrounded by distinct special tokens.
We introduce two generative approaches for encoder–decoder and decoder-only models, one of which---\textsc{Struct}---is essentially equivalent to \citet{bikaun-etal-2024-maintnorm}’s method.
Because few studies compare multiple generative approaches to LN---and none include decoder-only models---it is valuable to evaluate and contrast these methods across both architectures.

\begin{table}[t]
\centering
\footnotesize
\begin{tabular}{ll}
\toprule
Approach & Target Text \\
\midrule
\textsc{Plain} & \ja{\texttt{ツイッターみてる}} \\
\textsc{Struct} & \ja{\texttt{[[}ついった\texttt{>\,>}ツイッター\texttt{]]}みてる} \\
\textsc{Span} & \ja{ついった\texttt{>\,>}ツイッター\texttt{>\,>}\,0} \\
\bottomrule
\end{tabular}
\caption{Expected output text of each generative approach for an input text ``\ja{ついったみてる}.''}
\label{tab:gen_output}
\end{table}

As shown in Table~\ref{tab:gen_output}, the \textit{structured full-text} (\textsc{Struct}) approach generates a full normalized text with specifying the substrings before and after normalization and their spans, using symbols ``\texttt{[[},'' ``\texttt{>\,>},'' and ``\texttt{]]}.''
The other \textit{normalization span-only} (\textsc{Span}) approach generates not full-text but only substrings before and after normalization using a symbol ``\texttt{>\,>}.''
The number ($\geq 0$) succeeding the normalized substring represents how many times 
the same original substring occurred before that of interest in the original text, which is used for specifying the exact span of the original substring.
A symbol ``\texttt{||}'' is used to separate multiple normalization instances within the input text.
Only ``\texttt{NONE}'' is output when no normalization is necessary.

\paragraph{Encoder-Decoder}
The model is trained to generate the target text from each input source text via the standard sequence-to-sequence training.

\paragraph{Decoder-only}
The model takes as input a prompt with an instruction and a source text, like 
``Instruction:\,{\textbackslash}n\texttt{\{inst\}}{\textbackslash}n{\textbackslash}nInput:\,{\textbackslash}n\texttt{\{src\}}{\textbackslash}n{\textbackslash}nOutput:\,{\textbackslash}n'', 
and is trained to generate the target text ``\texttt{\{tgt\}} \texttt{EOS}'' via the standard instruction tuning. Here, \texttt{\{inst\}}, \texttt{\{src\}}, and \texttt{\{tgt\}} are placeholders, and \texttt{EOS} represents the end of text token.
We use English instruction texts explaining the corresponding content for \textsc{Struct} and \textsc{Span} as described above; the exact wording is provided in Appendix \ref{sec:prompt}.

\section{Experiments} \label{sec:exp}
We set the following experimental questions (EQs):
\begin{enumerate}
\setlength{\parskip}{0cm} 
\setlength{\itemsep}{0.1cm}
\item Across different model architectures, backbone models, and normalization approaches, which methods excel in normalization accuracy (precision and recall), and which methods are efficient in terms of inference cost?
\item How many training instances are required to achieve reasonable performance, and does this requirement vary by methods? 
\item What domains and other instance characteristics are particularly challenging to normalize?
\end{enumerate}
To address these EQs, we evaluated various LN methods using our Japanese dataset and an existing Thai dataset.
Specifically, experiments in \S\ref{sec:res_acc_ja}--\ref{sec:res_cost} (and \S\ref{sec:exp_detect}) correspond to EQ1, \S\ref{sec:train_size} to EQ2, and \S\ref{sec:res_domain} and \S\ref{sec:error_anal} to EQ3. 

We use precision, recall, and F${}_{0.5}$ score at the normalization-span level for individual non-standard words as evaluation metrics.
While normalized results with low recall do not harm downstream task performance compared to original texts, low-precision outputs introduce spurious tokens that can degrade downstream accuracy.
We therefore adopt the F${}_{0.5}$ score, which emphasizes precision over recall, similarly to other text correction tasks~\cite{yu-2021-gec}.
See Appendix~\ref{sec:def_metrics} for details of the score calculations.

\begin{table}[t]
\centering
\small
\begin{tabular}{cllrr}
\toprule
Lang & Dataset & Set & \#Sent & \#Norm\\
\midrule
\multirow{4}{*}{ja} & \multirow{4}{*}{JMLN} 
& train & 13,196 & 5,879 \\
&& dev & 1,880 & 791 \\
&& test${}^c$ & 3,786 & 1,705 \\
&& test${}^r$ & 2,540 & 510 \\
\midrule
\multirow{2}{*}{th} & \multirow{2}{*}{VISTEC-2021} 
& train & 40,000 & 130,790 \\
&& test & 10,000 & 32,819 \\
\bottomrule
\end{tabular}
\caption{Dataset statistics: The number of sentences (\#Sent) and normalization instances (\#Norm).}
\label{tab:stat}
\end{table}

\begin{table*}[t!]
\centering
\small
\begin{tabular}{clr|ccc|ccc|ccc}
\toprule
& Backbone & Size & \multicolumn{3}{c|}{\textsc{Full-Seg} Approach} & \multicolumn{3}{c|}{\textsc{Struct} Approach} & \multicolumn{3}{c}{\textsc{Span} Approach} \\
&&& P & R & F${}_{0.5}$ & P & R & F${}_{0.5}$ & P & R & F${}_{0.5}$\\
\midrule
E & BERT-base & 91M & 0.713 & \textbf{0.529} & 0.667 &--&--&--&--&--&--\\
E & RoBERTa-base & 100M & 0.718 & 0.523 & 0.669 &--&--&--&--&--&--\\
E & DeBERTa-base & 100M & \textbf{0.729} & 0.506 & \textbf{0.670} &--&--&--&--&--&--\\
\midrule
E & BERT-large & 310M & \textbf{0.762} & \textbf{0.570} & \textbf{0.714} &--&--&--&--&--&--\\
E & RoBERTa-large & 320M & 0.755 & 0.550 & 0.702 &--&--&--&--&--&--\\
E & DeBERTa-large & 330M & 0.750 & 0.568 & 0.705 &--&--&--&--&--&--\\
S & T5-base  & 250M &--&--&--& \underline{{0.701}} & 0.459 & 0.634 & 0.689 & \cellcolor{lblue}\underline{{0.508}} & \underline{{0.643}} \\ 
S & mT5-base & 580M &--&--&--& 0.606 & 0.406 & 0.551 & \cellcolor{llblue}\underline{0.655} & \cellcolor{llblue}\underline{0.427} & \cellcolor{llblue}\underline{0.592} \\ 
\midrule
S & T5-large & 780M &--&--&--& \underline{0.728} & 0.509 & \underline{0.670} & 0.704 & \underline{0.525} & 0.659 \\ 
S & mT5-large & 1.2B &--&--&--& 0.600 & 0.421 & 0.553 & \cellcolor{lblue}\underline{0.718} & \cellcolor{lblue}\underline{0.467} & \cellcolor{lblue}\underline{0.648} \\ 
D & Llama-3.2-1B    & 1.2B &--&--&--& \underline{0.654} & \underline{0.489} & \underline{0.612} & 0.626 & 0.480 & 0.590 \\ 
D & Sarashina2.2-1B & 1.4B &--&--&--& 0.722 & \underline{\textbf{0.612}} & 0.697 & \cellcolor{llblue}\underline{\textbf{0.761}} & 0.585 & \underline{\textbf{0.717}} \\ 
D & Qwen2.5-1.5B   & 1.5B &--&--&--& \underline{0.623} & {0.459} & \underline{0.580} & \cellcolor{pink}0.338 & \cellcolor{lblue}\underline{0.516} & \cellcolor{pink}0.363 \\ 
D & TinySwallow-1.5B& 1.5B &--&--&--& 0.605 & 0.556 & 0.593 & \cellcolor{lblue}\underline{0.667} & \underline{0.569} & \cellcolor{llblue}\underline{0.645} \\ 
\midrule
D & Qwen2.5-3B     & 3.1B &--&--&--& \underline{0.598} & \underline{0.537} & \underline{0.583} & \cellcolor{pink}0.376 & \underline{0.537} & \cellcolor{pink}0.400 \\ 
D & Llama-3.2-3B    & 3.2B &--&--&--& \underline{0.680} & 0.522 & \underline{0.641} & 0.666 & \underline{0.530} & 0.633 \\ 
D & Sarashina2.2-3B & 3.4B &--&--&--& 0.774 & \underline{\textbf{0.668}} & 0.751 & \underline{\textbf{0.781}} & 0.660 & \underline{\textbf{0.754}} \\
\midrule
D & Sarashina2-7B    & 7.3B &--&--&--& 0.743 & 0.649 & 0.722 & \underline{0.744} & \underline{\textbf{0.657}} & \underline{\textbf{0.724}} \\ 
D & Qwen2.5-7B      & 7.6B &--&--&--& \underline{0.717} & \underline{0.558} & \underline{0.678} & \cellcolor{pink}0.379 & \cellcolor{llblue}0.589 & \cellcolor{pink}0.408 \\ 
D & Llama-3.1-8B     & 8.0B &--&--&--& \underline{0.738} & 0.538 & \underline{0.687} & 0.719 & \underline{0.549} & 0.677 \\ 
D & $\hookrightarrow$Swallow-8B & 8.0B &--&--&--& \underline{\textbf{0.749}} & \underline{0.605} & \underline{0.715} & 0.741 & 0.602 & 0.708 \\
\bottomrule
\end{tabular}
\caption{JMLN test results of Japanese LN models (E: encoder, S: seq2seq, D: decoder). ``$\hookrightarrow$'' indicates the continual pre-trained model derived from the base model listed in the previous row. 
For each size group (separated by horizontal lines), the best score is indicated in \textbf{bold}. For each backbone model, the better of the \textsc{Struct} and \textsc{Span} approaches is \underline{underlined}. Scores where the \textsc{Span} approach shows a +5\%, +10\%, or -10\% increase/decrease compared to the \textsc{Struct} approach are highlighted with 
\textcolor{lblue}{$\blacksquare$}\,blue, \textcolor{llblue}{$\blacksquare$}\,light blue, and \textcolor{pink}{$\blacksquare$}\,pink backgrounds, respectively.
}
\label{tab:res_ja}
\end{table*}

\subsection{Datasets} \label{sec:exp_dataset}
Table~\ref{tab:stat} shows the statistics of two experimental datasets: JMLN and Thai VISTEC-2021 \cite{limkonchotiwat-etal-2021-handling}.\footnote{\url{https://github.com/mrpeerat/OSKut/tree/main/VISTEC-TP-TH-2021}} 
For JMLN, we divided the annotation data for each domain (01--14) into train, dev, and test${}^c$ sets, and merged the all train, dev, and test${}^c$ sets into unified train, dev, and test${}^c$ sets across all domains, respectively.\footnote{We also created an additional test set, the test${}^r$ set, but focus on experiments with the test${}^c$ set. See
Appendix~\ref{sec:data_creation_detail} for more details on the test${}^r$ set.}
Unless otherwise stated, we simply refer to the test${}^c$ set as the ``test set.''
For VISTEC-2021, we followed the provided training/test split and regarded randomly-sampled 5\% sentences in the training set as a dev set. 

\subsection{Models}
As the backbone of LN systems, we used the following Japanese or multilingual pre-trained models in Japanese experiments:
BERT~\cite{devlin-etal-2019-bert}, RoBERTa~\cite{liu-2019-roberta}, and DeBERTa~\cite{he2021deberta} as character-level encoder-only models,
T5~\cite{raffel-etal-2020-t5} and mT5~\cite{xue-etal-2021-mt5} as encoder-decoder models, and Llama 3.1/3.2~\cite{grattafiori-2024-llama3}, Qwen2.5~\cite{qwen-2025}, Lllama-3.1-Swallow~\cite{fujii-2024-continual}, TinySwallow~\cite{shing-2025-taid} and Sarashina2/2.2~\cite{sbintuitions-sarashina2,sbintuitions-sarashina22} as decoder-only models.\footnote{Languages supported by each model instance are shown in Table~\ref{tab:backbone}. Notably, the encoder-only models used in this study tokenize at the character level, whereas the other models employ subword-level tokenization.}

Similarly, we used the following Thai or multilingual 
pre-trained models in the Thai experiments: RoBERTa~\cite{yasuoka-2023-sequence} as a character-level encoder-only model, T5 and mT5 as encoder-decoder models, and Llama 3.1/3.2, Qwen2.5, Typhoon 2~\cite{typhoon2}, OpenThaiGPT 1.5~\cite{yuenyong-2025-openthaigpt}, and SeaLLMs 3~\cite{zhang-2024-seallms3} as decoder-only models.
Specific model instances are listed in Appendix~\ref{sec:pretrained_models}.

We fine-tuned each model, with
applying LoRA~\cite{hu2022lora} or QLoRA~\cite{dettmers2023qlora} to some decoder-only models, and selected the model checkpoint with the best F${}_{0.5}$ score on the dev set.
We fine-tuned all models twice with different random seeds and report mean scores for two runs.
The hyperparameter settings are listed in Appendix \ref{sec:hypara}.

\section{Results and Analysis}
\subsection{Normalization Accuracy for Japanese} \label{sec:res_acc_ja}
We evaluated LN methods with three types of architectures on the JMLN test set.
Table~\ref{tab:res_ja} shows the performance of encoder-only models, and encoder-decoder and decoder-only models with both \textsc{Struct} and \textsc{Span} approaches.\footnote{The \textsc{Plain} approach yielded performance similar to the other two approaches on the dev set, as shown in \S\ref{sec:anal_plain}.}

The observed results are as follows.
(1) Among encoder-only models, the large models outperformed base models, while different backbone models showed similar performance. 
(2) Among generative methods, the \textsc{Span} approach basically achieved performance comparable to or better than the \textsc{Struct} approach in many cases.\footnote{The only exception is the Qwen2.5 model series; the models generated many nonsensical text flagments---such as ``str1\texttt{>\,>}str2\texttt{>\,>}0,'' where ``str1'' did not appear in the original input text---until reaching the maximum output length, resulted in low precision and long inference time.} 
(3) Within the same model series---T5, mT5, Llama-3.2, Qwen2.5, and Sarashina2.2---performance improved with increasing model size up to 8B (see Figure in Appendix \ref{sec:model_series_figure}). However, in our preliminary experiment on the dev set, we observed no salient additional gains from the 13B--15B models.
(4) Performance within groups of similarly sized models was not equivalent; certain series---specifically Sarashina2.2 series, followed by Swallow, demonstrated saliently superior performance.
(5) These strong decoder-only models outperformed models with the other two architectures in recall.

In conclusion, encoder-only models demonstrated high performance despite its small size, and Sarashina2.2-3B model achieved the highest performance overall, indicating that the high capability of the backbone model was beneficial for this task.

\subsection{Normalization Accuracy for Thai} \label{sec:res_acc_th}
On the Thai VISTEC test set, we evaluated LN methods, with only the \textsc{Span} approach for generative methods.
Table~\ref{tab:res_th} shows the results.

\begin{table}[t]
\centering
\small
\begin{tabular}{clr|ccc}
\toprule
& Backbone & Size & P & R & F${}_{0.5}$ \\
\midrule
E & RoBERTa-base & 88M & \textbf{0.713} & \textbf{0.529} & \textbf{0.666} \\
\midrule
S & T5-base  & 250M & \textbf{0.645} & 0.618 & \textbf{0.640} \\
S & mT5-base & 580M & 0.642 & \textbf{0.629} & 0.639 \\
\midrule
S & mT5-large & 1.2B & \textbf{0.660} & 0.609 & \textbf{0.649} \\
D & Llama-3.2       & 1.2B & 0.628 & 0.628 & 0.628 \\
D & $\hookrightarrow$ Typhoon2 & 1.2B & 0.644 & 0.634 & 0.642 \\
D & SeaLLMs3 & 1.5B & 0.462 & {0.689} & 0.495 \\
D & Qwen2.5   & 1.5B & 0.472 & \textbf{0.702} & 0.505 \\
\midrule
D & Qwen2.5     & 3.1B & 0.457 & \textbf{0.706} & 0.492 \\
D & Llama-3.2    & 3.2B & 0.641 & 0.647 & 0.642 \\
D & $\hookrightarrow$ Typhoon2 & 3.2B & \textbf{0.656} & 0.668 & \textbf{0.658} \\
\midrule
D & SeaLLMs3    & 7.6B & 0.465 & 0.705 & 0.499  \\
D & Qwen2.5      & 7.6B & 0.461 & \textbf{0.709} & 0.496 \\
D & $\hookrightarrow$ ThaiGPT1.5 & 7.6B & 0.653 & 0.672 & 0.657 \\
D & Llama-3.1     & 8.0B & 0.653 & 0.659 & 0.655 \\
D & $\hookrightarrow$ Typhoon2 & 8.0B & \textbf{0.661} & 0.678 & \textbf{0.664} \\
\bottomrule
\end{tabular}
\caption{VISTEC test results of Thai LN models.}
\label{tab:res_th}
\end{table}

Similarly to the Japanese results, performance improved with increasing model size within the same model series.
Additionally, continually pre-trained models focusing on Thai outperformed their base models.
Overall, the small encoder-only RoBERTa-base exhibited the best precision, while all encoder-decoder and decoder-only models surpass it in recall.
This introduces a precision-recall trade-off in model selection.

\subsection{Inference Throughput} \label{sec:res_cost}
For selected models with high normalization accuracy, we measured their inference throughput using the JMLN test set, on both an NVIDIA V100 GPU with 32 GiB memory and an H200 GPU with 140 GiB memory (see detailed settings in Appendix \ref{sec:setting_cost}).
Table~\ref{tab:cost} shows the results.

\begin{table}[t]
\centering
\small
\begin{tabular}{l|rr|rr}
\toprule
Model & V100 & H200  & V100 & H200 \\
\midrule
& \multicolumn{2}{c|}{\textsc{Full-Seg}} & \\
\midrule
BERT-large    & 508.8 & 1420.9 & -- & -- \\
RoBERTa-large & 561.4 & 1407.2 & -- & -- \\
DeBERTa-large & 395.0 & 1038.3 & -- & -- \\
\midrule
& \multicolumn{2}{c|}{\textsc{Struct}} & \multicolumn{2}{c}{\textsc{Span}} \\
\midrule
T5-large        & 136.0 & 312.3 & 159.0 & 208.4\\
Sarashina2.2-1B & 66.5  & 202.1 & 75.4  & 243.1\\
Sarashina2.2-3B & 32.6  & 118.1 & 36.0  & 117.4\\
Sarashina2-7B   & 17.8  &  73.3 & 19.8  &  83.4 \\
\bottomrule
\end{tabular}
\caption{Throughput: the number of sentences processed per second, measured on a V100 and H200 GPU.}
\label{tab:cost}
\end{table}

\begin{figure}[t]
\centering
\includegraphics[width=\linewidth]{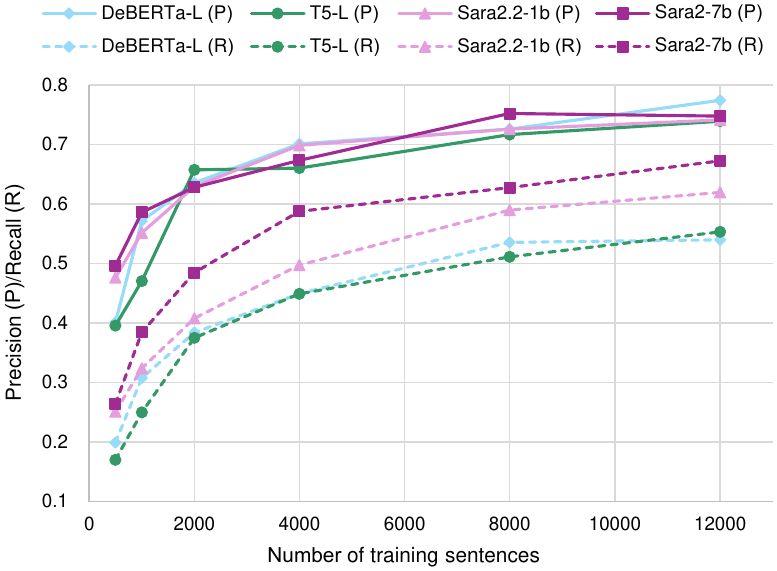}
\caption{JMLN test results for each training data size.}
\label{fig:res_size}
\end{figure}

We observed:
(1) Encoder-only models were the fastest, followed by T5, the smaller Sarashina model, and finally the larger Sarashina model.
(2) The \textsc{Span} approach yielded modest gains over \textsc{Struct}, except for T5 on the H200. Since instruction text occupies a large proportion of total output tokens, more concise instruction prompts can improve throughput; however, it is necessary to explore prompts that preserve normalization accuracy.
(3) Sarashina models exhibited substantially lower throughput on the V100 than on the H200. Their low throughput on the V100 is a critical drawback, but they run much faster on the H200. Thus, when high-spec GPUs are available, Sarashina models are viable options in accuracy-critical scenarios.

\begin{table*}[t]
\centering
\footnotesize
\begin{tabular}{ll|c|c|cc|cc|cc}
\toprule
Data  & Register & Surf-Out-Train & Avg-3M & \multicolumn{2}{c|}{DeBERTa-L} & \multicolumn{2}{c|}{T5-L} & \multicolumn{2}{c}{Sarashina2.2-3B} \\
&&& F${}_{0.5}$ & P & R & P & R & P & R \\
\midrule
01~~\texttt{BJ-OC}  & Q\&A site     & \underline{0.59} & \underline{0.595} & 0.654 & 0.416 & 0.572 & 0.369 & \textbf{0.717} & \textbf{0.578} \\
02~~\texttt{BJ-OY}  & Blog          & 0.48 & 0.713 & 0.720 & 0.543 & 0.741 & 0.551 & \textbf{0.802} & \textbf{0.662} \\
03~~\texttt{RC-BLG} & Blog          & 0.37 & 0.775 & 0.801 & 0.691 & 0.793 & 0.664 & \textbf{0.803} & \textbf{0.728} \\
04~~\texttt{RC-REV} & Reviews       & 0.43 & 0.790 & \textbf{0.868} & 0.731 & 0.790 & 0.593 & 0.808 & \textbf{0.741} \\
05~~\texttt{RK-ICB} & Reviews       & \underline{0.52} & \underline{0.680} & \textbf{0.808} & 0.458 & 0.693 & 0.500 & 0.764 & \textbf{0.521} \\
06~~\texttt{RK-TRV} & Reviews       & 0.40 & 0.715 & 0.763 & 0.490 & \textbf{0.820} & 0.433 & {0.795} & \textbf{0.663} \\
07~~\texttt{RK-RCP} & Recipes       & 0.35 & 0.792 & 0.815 & 0.684 & 0.817 & 0.663 & \textbf{0.834} & \textbf{0.734} \\
08~~\texttt{AM}     & Reviews       & 0.32 & 0.798 & \textbf{0.862} & 0.728 & 0.776 & 0.655 & 0.832 & \textbf{0.748} \\
09~~\texttt{NC-VID} & Video desc.   & 0.39 & \underline{0.645} & 0.627 & 0.538 & 0.666 & 0.492 & \textbf{0.730} & \textbf{0.629} \\
10~~\texttt{NC-PED} & Forum         & 0.27 & 0.795 & \textbf{0.850} & 0.747 & 0.783 & 0.677 & 0.808 & \textbf{0.772} \\
11~~\texttt{TW}     & Social media  & \underline{0.60} & \underline{0.633} & 0.613 & 0.511 & 0.600 & 0.464 & \textbf{0.763} & \textbf{0.677} \\
12~~\texttt{JW}     & Wiki hist. & \underline{0.69} & \underline{0.312} & \textbf{0.528} & 0.140 & 0.197 & 0.095 & 0.509 & \textbf{0.280} \\
13~~\texttt{NU}     & Conv. trans. & 0.17 & 0.758 & \textbf{0.845} & \textbf{0.745} & 0.683 & 0.596 & 0.800 & \textbf{0.745} \\
14~~\texttt{SK}     & Conv. trans. & 0.43 & 0.718 & 0.771 & 0.550 & 0.728 & 0.457 & \textbf{0.830} & \textbf{0.664} \\
\midrule
 All & & 0.44 & 0.706 & 0.750 & 0.568 & 0.704 & 0.525 & \textbf{0.781} & \textbf{0.660} \\
\bottomrule
\end{tabular}
\caption{JMLN test results of representative three models for each domain. ``Surf-Out-Train'' indicates the Surf-Outside-Train rate for each domain test set. Avg-3M indicates the average of F${}_{0.5}$ scores of the three models.
(Surf-Outiside-Train rate values above 0.5 and Avg-3M values below 0.7 are highlighted by \underline{underlining}.)} \label{tab:res_ja_domain}
\end{table*}

\begin{table*}[t]
\centering
\small
\begin{tabular}{ll|ccc|ccc|cc}
\toprule
Backbone & Approach & \multicolumn{3}{c|}{Det} & \multicolumn{3}{c|}{Norm} & \multicolumn{2}{c}{Det$-$Norm}\\
&& P & R & F${}_{0.5}$ & P & R & F${}_{0.5}$ & P & R \\
\midrule
DeBERTa-large & \textsc{Full-Seg} & \textbf{0.865} & 0.655 & \textbf{0.813} & 0.750 & 0.568 & 0.705 & 0.115 & 0.087 \\
T5-large & \textsc{Span} & 0.801 & 0.597 & 0.750 & 0.704 & 0.525 & 0.659 & 0.097 & 0.072 \\
Sarashina2.2-3B & \textsc{Span} & 0.829 & \textbf{0.700} & 0.800 & \textbf{0.781} & \textbf{0.660} & \textbf{0.754} & 0.048 & 0.041 \\
\bottomrule
\end{tabular}
\caption{JMLN test results of representative three models for detection (Det) and normalization (Norm) tasks. The best score among three models is highlighted in \textbf{bold}.}
\label{tab:res_ja_det}
\end{table*}

\subsection{Investigation of Training Data Size} \label{sec:train_size}
We generated size-$N$ training sets by sampling random $N$\,$\in$\,$\{500, 1\mbox{k}, 2\mbox{k}, 4\mbox{k}, 8\mbox{k}, 12\mbox{k}\}$  sentences from the entire JMLN training set, and we then fine-tuned each of DeBERTa, T5, and Sarashina models twice for each size-$N$ training set.
As shown in Figure~\ref{fig:res_size}, the results are as follows.

First, a general trend across all models is that precision and recall improve as the data size increases. From sizes 8k to 12k the gains are more gradual, but performance is not yet saturated. Within the evaluated range, more data yields better results; however, even a 4k to 8k-size dataset can achieve reasonable precision around 0.70 when creating large amounts of annotated data is impractical.  

Second, in model‐specific comparisons, precision and recall follow different patterns. Precision shows no clear differences across models. In contrast, recall is consistently highest for Sarashina2-7B, followed by Sarashina2.2-1B, and lower for DeBERTa and T5; this indicates that the Sarashina-series models generalize well in terms of coverage, even when the training data size is small.  

\subsection{Results Across Domains} \label{sec:res_domain}
As shown in Table~\ref{tab:res_ja_domain}, we evaluated the performance of selected models---DeBERTa-large, T5-large (\textsc{Span}), and Sarashina-2.2-3B (\textsc{Span})---on each domain test set of JMLN (results on grouped domains are provided in Appendix \ref{sec:res_domain_agg}).

First, to explore what makes a domain difficult, we examined an indicator: 
the proportion of non-standard surface tokens in the test set that are not found among the training set’s non-standard surfaces (Surf-Outside-Train rate).
We then computed the Pearson correlation coefficients $r$ between the indicator and the average F${}_{0.5}$ scores across the three models, obtaining a strong negative correlation ($r=-0.78$).
Notably, for 9 out of 10 domains with a Surf-Outside-Train rate below 0.5 had average F${}_{0.5}$ scores above 0.7, whereas all 4 domains with a rate above 0.5 had average scores below 0.7.

Next, model performance comparisons revealed the following.
(1) In all domains, all models exhibited higher precision than recall, showing a desirable characteristic because invalid normalizations would degrade downstream task performance.
(2) Across most domains, Sarashina-2.2 achieved higher recall than the other models, resulting its superior overall performance. Notably, this model achieved recall over 0.5 across typical UGT domains (01--11).
(3) 
All models exhibited very low recall below 0.3 in domain 11---a specialized domain data originated from \citeauthor{tanaka-etal-2020-building} (\citeyear{tanaka-etal-2020-building})'s typo correction dataset. Training with the task-specific dataset would improve performance, but we leave this for future.

\subsection{Informal Span Detection Accuracy} \label{sec:exp_detect}
We evaluated the performance of the informal word span detection task for the same three Japanese models as in \S\ref{sec:res_domain}.\footnote{We report the detection accuracy of the Thai models in Appendix~\ref{sec:det_thai}.}
In this task, a model’s prediction is regarded as correct if the predicted span of a non-standard word matches the gold standard, irrespective of the predicted standard form.
Table~\ref{tab:res_ja_det} shows the detection results, the normalization results, and the precision and recall differences between the two tasks.

DeBERTa achieved the best precision and F${}_{0.5}$ score among the models for the detection task, but its precision and recall dropped by about 0.09--0.12 for the normalization task.
Sarashina2.2 achieved the best recall for the detection task, with smaller drops of about 0.04--0.05 precision and recall for the normalization task, resulting in the best overall normalization performance.
T5 yielded the worst results; the precision/recall differences between the two tasks were intermediate between those of the remaining models.
These results suggest that DeBERTa and T5 leave notable room for improvement in generating accurate standard forms, whereas the best-performing Sarashina2.2 shows only a minor gap. Thus, further accuracy gains will likely depend on better handling of cases where span detection fails.

\subsection{Error Analysis} \label{sec:error_anal}
For the three models evaluated in \S\ref{sec:res_domain} and \S\ref{sec:exp_detect}, we analyzed output patterns on the JMLN dev set.
Specifically, we counted predictions, true positives (TP), false positives (FP), and false negatives (FN) for each model, and measured (i) the proportion of predicted normalized forms appearing in the UniDic lexicon (Norm-In-Lex rate) and (ii) the proportion of original surface forms for TP, FP, and FN instances that matched any non-standard forms in the training set (Surf-In-Train rate).
Table~\ref{tab:error_stat} shows the results averaged over two runs per model.

DeBERTa exhibited a notably lower Norm-In-Lex rate than the other models (0.921--0.928) and the gold standard (0.947).
By manually inspecting error cases, we found that DeBERTa's restored tokens within spans sometimes formed nonsensical words.\footnote{E.g., pred. \ja{かど}, orig. \ja{ヶド}, gold \ja{けど}/\ja{けれど} (``but'') and pred. \ja{だじりん}, orig. \ja{だぁりん}, gold \ja{ダーリン} (``darling'').}
Both suggest that the model's independent prediction at each \texttt{MASK} position makes it especially prone to such errors.

Regarding the Surf-In-Train rate, all models exhibited similar trends. For TPs, approximately 70--80\% of the original surface forms were known (i.e., appeared in the training set), indicating that many correct predictions relied on the seen normalization instances. For both FPs and FNs, only approximately 20--30\% were known, indicating that the majority of errors involved unseen expressions. This suggests considerable room for improving generalization in normalizing unseen cases.
These results align with the findings in \S\ref{sec:res_domain}.
For actual prediction examples, see Appendix \ref{sec:error_examples}.

\begin{table}[t]
\centering
\small
\begin{tabular}{l|ccc}
\toprule
& DeBERTa & T5 & Sarashina2.2\! \\
\midrule
\#Predictions            & 614.5 & 598.5 & 676.5 \\
Norm-In-Lex rate       & 0.875 & 0.928 & 0.921 \\
\midrule
\#TPs                    & 469.0 & 449.0 & 535.5 \\
Surf-In-Train rate     & 0.806 & 0.806 & 0.716 \\
\midrule
\#FPs                    & 145.5 & 149.5 & 141.0 \\
Surf-In-Train rate      & 0.289 & 0.183 & 0.238 \\
\midrule
\#FNs                    & 322.0 & 342.0 & 255.5 \\
Surf-In-Train rate      & 0.258 & 0.290 & 0.302 \\
\bottomrule
\end{tabular}
\caption{Models' prediction statistics on JLMN dev set (DeBERTa: large, T5: large, Sarashina2.2: 3B).}
\label{tab:error_stat}
\end{table}

\section{Conclusion}
This paper presented our multi-domain Japanese LN dataset, LN methods based on three Transformer architectures for unsegmented languages, and multi-perspective experiments and analysis.

The answers to the three evaluation questions (\S\ref{sec:exp}) are summarized as follows. 
(1) Compact encoder-only models achieved high precision and offered the best inference throughput, while cutting-edge decoder-only models delivered high precision, notably high recall, and reasonable throughput on a high-spec GPU.
(2) Normalization accuracy consistently increased with training data size, yet even 4k--8k training sentences yielded reasonable precision around 0.7. Sarashina-series models, in particular, achieved superior recall with fewer training sentences.
(3) Domains with higher rates of unknown non-standard tokens correlated with decreased performance across models. Typo correction emerged as the most challenging category, reflecting the difficulty posed by diverse typo patterns.

In future work, we will evaluate the impact of LN on downstream tasks and explore the development of a general-purpose decoder model with robust normalization capabilities.

\section*{Limitations}
\paragraph{Generalizability of Experimental Results}
Our findings and conclusions are presented in the context of the datasets, languages, and models used in this study.

\paragraph{Dataset Size}
The experimental results in \S\ref{sec:res_acc_ja}---high precision up to 0.78---indicates that our dataset is large enough to train high-accuracy models. However, results in \S\ref{sec:train_size} show no clear saturation even at the maximum training size, suggesting that additional data could further improve performance. Given the cost limits of manual annotation, a promising direction is to explore methods for generating high-quality synthetic data.

\paragraph{Language Coverage}
We evaluated our methods only on Japanese and Thai datasets, but they are readily applicable to other unsegmented and space-delimited languages. Validation on additional languages remains future work.

\paragraph{Inference Throughput Settings}
To ensure fair comparison, we measured throughput using a single GPU via the Hugging Face Transformers~\cite{wolf-etal-2020-transformers} library. However, throughput could be improved through multi-GPU parallelism, model quantization, or adoption of high-performance inference engine, such as vLLM~\cite{kwon2023vllm}.

\paragraph{Encoder-Only Architecture Variants}
A state-of-the-art encoder-only model, ModernBERT~\cite{warner2024modernbert}, might achieve performance on par with or exceeding the models we evaluated. Due to computational and time constraints, this remains future investigation.

\paragraph{Word-Level Evaluation Metrics}
Our word-level normalization metrics treat any span mismatch as an error---even if the predicted normalization is semantically valid---which we observed especially in decoder-only model outputs (see Appendix \ref{sec:error_examples} for examples). Such span differences have little impact on most downstream applications, so the practical usefulness of the outputs may exceed the scores reported. To complement word-level metrics, we also provide sentence-level exact-match accuracy and the chrF score~\cite{popovic-2015-chrF} in Appendix \ref{sec:anal_enc} and \ref{sec:anal_plain}.

\section*{Ethics Statement}
\paragraph{License of Resources}
MeCab is available under GPL, LGPL, and BSD License.
UniDic (\texttt{unidic-cwj-3.1.0}) is available under GPL v2.0, LGPL v2.0, and New BSD License.
sacreBLEU is available under Apache License 2.0 (we used this software for preliminary experiments shown in Appendix \ref{sec:detailed_exp}).
The licenses for the datasets and pre-trained models are listed in Appendices \ref{sec:pretrained_models} and \ref{sec:data_src} (Table~\ref{tab:backbone}).
Our use of these resources for academic research aligns with their intended use.
We will release our JMLN dataset for academic research in information science; it will include only annotation information and not the original texts.\footnote{The original texts have already been released by the original dataset providers, and we will publish our annotation dataset along with a tool that integrates those texts with the annotations, enabling users to reconstruct the complete dataset.}

\paragraph{Human Annotators}
The annotation work was performed by annotators at a professional data annotation company.
The payment amount to the company was based on the estimate submitted by the company.
The actual annotators and the payment amount to each annotator were determined by the company.
The annotation work was performed by four annotators, including an annotation manager, all of whom are native Japanese speakers.
Under the contract for the annotation work, it was agreed that the intellectual property rights to the deliverables would be transferred to the authors' institution.

\paragraph{Potential Risks}
Appropriate normalization can facilitate NLP applications while preserving the core meaning of the original texts. However, it may diminish subtle nuances and intentions in the original text; for example, casual expressions may be rendered formal, dialectal expressions may be replaced by semantically similar standard language forms, or incorrect normalization may produce an entirely different meaning.

\section*{Acknowledgments}
We would like to thank the anonymous reviewers and meta reviewers at ACL Rolling Review for their constructive feedback, especially the reviewer whose detailed comments and extensive suggestions were helpful in improving the quality of this paper.
To construct our dataset, we made use of existing datasets: the Balanced Corpus of Contemporary Written Japanese, the Rakuten Dataset, the Multilingual Amazon Reviews Corpus, the Japanese Wikipedia Typo Dataset, the Nagoya University Conversation Corpus, and the Japanese and Chinese Skype Conversation Corpus, as well as the Recruit Dataset provided by Recruit Co., Ltd. and the Niconico Dataset provided by DWANGO Co., Ltd., both through the IDR Dataset Service of the National Institute of Informatics.

\bibliography{mybib}

\clearpage
\appendix
\section{JMLN Dataset}
\subsection{Details on Data Creation} \label{sec:data_creation_detail}
Before conducting the sentence selection step described in \S\ref{sec:data_const}, we divided each candidate set from each of the original 14 datasets into two subsets: a manually curated (Cur) set and a random (Rand) set.
The successive steps for the Cur set are those described in \S\ref{sec:data_const}, and the resulting annotation data are divided into train, dev, and test${}^c$ sets, as in \S\ref{sec:exp_dataset}.

For the Rand set, annotators sequentially selected sentences from candidates of each original dataset, provided that they did not contain ethically problematic contents or unclear meaning. We consider the entire resulting annotation data as an additional test set, which we refer to as the test${}^r$ set.
This dataset is considered to exhibit a distribution of non-standard word frequencies close to that of natural text, making it useful for evaluations that assume such data. However, experiments using this dataset are left for future work.

\subsection{Data Statistics} \label{sec:data_stat}
The detailed dataset statistics of JMLN are shown in Table~\ref{tab:stat_mset} and Table~\ref{tab:stat_rset}.
Note that sentences in the Cur set are divided into train, dev, and test${}^c$ sets, and the Rand set corresponds to the test${}^r$ set.

\begin{table}[h]
\centering
\footnotesize
\begin{tabular}{llrrr}
\toprule
ID & Name & \#Sent & \#Word & \#Norm \\
\midrule
01 & \texttt{BJ-OC} & 1,441 & 28,631 & 928 \\
02 & \texttt{BJ-OY} & 1,785 & 29,092 & 1,445 \\
03 & \texttt{RC-BLG} & 2,312 & 37,135 & 766 \\
04 & \texttt{RC-REV} & 1,541 & 31,283 & 310 \\
05 & \texttt{RK-ICB} & 1,251 & 21,548 & 251 \\
06 & \texttt{RK-TRV} & 1,610 & 29,240 & 289 \\
07 & \texttt{RK-RCP} & 2,479 & 29,834 & 1,104 \\
08 & \texttt{AM} & 1,769 & 28,055 & 477 \\
09 & \texttt{NC-VID} & 918 & 11,908 & 262 \\
10 & \texttt{NC-PED} & 947 & 14,858 & 387 \\
11 & \texttt{TW} & 1078 & 14,701 & 858 \\
12 & \texttt{JW} & 540 & 18,817 & 503 \\
13 & \texttt{NU} & 507 & 9,176 & 509 \\
14 & \texttt{SK} & 684 & 11,935 & 286 \\
\midrule
& Total & 18,862 & 316,216 & 8,375 \\ 
\bottomrule
\end{tabular}
\caption{Statistics of the JMLN Cur set.} \label{tab:stat_mset}
\end{table}
\vspace{-1pc}

\begin{table}[h]
\centering
\footnotesize
\begin{tabular}{llrrr}
\toprule
ID & Name & \#Sent & \#Word & \#Nrom \\
\midrule
01 & \texttt{BJ-OC} & 200 & 3,981 & 33 \\
02 & \texttt{BJ-OY} & 201 & 3,821 & 56 \\
03 & \texttt{RC-BLG} & 200 & 2,903 & 57 \\
04 & \texttt{RC-REV} & 200 & 3,872 & 27 \\
05 & \texttt{RK-ICB} & 200 & 2,942 & 16 \\
06 & \texttt{RK-TRV} & 200 & 3,139 & 18 \\
07 & \texttt{RK-RCP} & 200 & 2,763 & 25 \\
08 & \texttt{AM} & 200 & 2,932 & 12 \\
09 & \texttt{NC-VID} & 150 & 2,418 & 23 \\
10 & \texttt{NC-PED} & 182 & 2,672 & 46 \\
11 & \texttt{TW} & 207 & 3,127 & 99 \\
13 & \texttt{NU} & 200 & 2,940 & 82 \\
14 & \texttt{SK} & 200 & 2,182 & 16 \\
\midrule
& Total & 2,540 & 39,692 & 510 \\
\bottomrule
\end{tabular}
\caption{Statistics of the JMLN Rand set.} \label{tab:stat_rset}
\end{table}

\subsection{Data Sources and Licenses} \label{sec:data_src}

To construct our JMLN dataset, we used following datasets and text sources as shown in Table~\ref{tab:data_src}: 
\begin{itemize}
\setlength{\parskip}{0cm} 
\setlength{\itemsep}{0.1cm}
\item (01--02) Balanced Corpus of Contemporary Written Japanese (BCCWJ) \cite{maekawa2014}: available under a usage contract;\footnote{\url{https://clrd.ninjal.ac.jp/bccwj/en/index.html}}
\item (03--04) Recruit Dataset~\cite{recruit_data}: available under a usage contract;\footnote{\url{https://www.nii.ac.jp/dsc/idr/recruit/}}
\item (05--07) Rakuten Dataset: available under a usage contract;\footnote{\url{https://rit.rakuten.com/data_release/}}
\item (08) Multilingual Amazon Reviews Corpus \cite{keung-etal-2020-multilingual}: previously available under a proprietary license (now unavailable);\footnote{\url{https://registry.opendata.aws/amazon-reviews-ml/}} 
\item (09--10) Niconico Dataset~\cite{nicovideo_data,nicopedia_data}: available under specific terms of use;\footnote{\url{https://www.nii.ac.jp/dsc/idr/nico/}}
\item (11) Twitter (now X) posts: obtained via the Twitter streaming API (copyright retained by each post's author);
\item (12) Japanese Wikipedia Typo Dataset \cite{tanaka-etal-2020-building}: available under CC-BY-SA 3.0 license;\footnote{\url{https://nlp.ist.i.kyoto-u.ac.jp/EN/?JWTD}}
\item (13) Nagoya University Conversation Corpus \cite{fujimiya-etal-2012}: available under CC BY-NC-ND 4.0 license;\footnote{\url{https://mmsrv.ninjal.ac.jp/nucc/}}
\item (14) Japanese and Chinese Skype Conversation Corpus: available under specific terms of use.\footnote{\url{http://nakamata.info/database/}}
\end{itemize}

\begin{table*}[t]
\centering
\footnotesize
\begin{tabular}{llllr}
\toprule
ID & Data name & Source dataset & Text register & \multicolumn{1}{l}{Year} \\
\midrule
01 & \texttt{BJ-OC} & BCCWJ: Yahoo! Chiebukuro & Q\&A posts/responces & 2004--2005 \\
02 & \texttt{BJ-OY} & BCCWJ: Yahoo! Blog & Blog posts & 2008--2009 \\
03 & \texttt{RC-BLG} & Recruit: beauty salon blogs & Blog posts & 2012--2014 \\
04 & \texttt{RC-REV} & Recruit: beauty salon reviews & Reviews (beauty salon) & 2012--2014\\
05 & \texttt{RK-ICB} & Rakuten Ichiba & Reviews (EC site) & 2019 \\
06 & \texttt{RK-TRV} & Rakuten Travel & Reviews (hotel site) & 2017--2019 \\
07 & \texttt{RK-RCP} & Rakuten Recipe & Recipes & 2017 \\
08 & \texttt{AM} & Multilingual Amazon Reviews Corpus & Reviews (EC site) & 2000--2015 \\
09 & \texttt{NC-VID} & Niconico Dataset: Video meta data & Video descriptions & 2018 \\
10 & \texttt{NC-PED} & Niconico Dataset: Forum data & Forum posts/replies & 2008--2014 \\
11 & \texttt{TW} & Twitter & Social media posts& 2020--2022 \\
12 & \texttt{JW} & Japanese Wikipedia Typo Dataset & Encyclopedia edit history & --2021 \\
13 & \texttt{NU} & Nagoya University Conversation Corpus & Conv. transcriptions & 2001--2003 \\
14 & \texttt{SK} & Skype Conversation Corpus & Conv. transcriptions & 2012 \\
\bottomrule
\end{tabular}
\caption{Data sources of JMLN. The Year column indicates the years of original text publication.} \label{tab:data_src}
\end{table*}

\begin{table*}[h!]
\centering
\footnotesize
\begin{tabular}{lllll}
\toprule
\multicolumn{2}{c}{} & Example  & Standard forms & Translation \\
\midrule
\multirow{10}{*}{Vocabulary type}
& Neologisms/Slang & \ja{コピペ}   & -- & copy and paste\\
& Proper names     & \ja{ドラクエ} & -- & Dragon Quest \\
& Onomatopoeia     & \ja{キラキラ} & -- & glitter \\
& Interjections    & \ja{おお}    & -- & oops \\ 
& Dialect words    & \ja{ほんま}  & -- & truly \\
& Foreign words    & \ja{ＥＡＳＹ} & -- & easy \\
& Ancient words*   & \ja{[行く] べし}    & -- & should [go]\\ 
& Character endings*& \ja{[行く] にゃ}   & -- & $\dagger$\\ 
& Blend words*     & \ja{おはこんばんにちは} & -- & $\ddagger$ \\ 
& Emoticons/AA     & \ja{（＾−＾）} & -- \\
\midrule
\multirow{4}{*}{Variant-form type}
& Character type variants  & \ja{カワイイ}   & \ja{かわいい,可愛い} & cute \\
& Alternative representations & \ja{大きぃ}  & \ja{大きい}  & big \\
& Sound change variants    & \ja{おいしーい} & \ja{おいしい,美味しい} & tasty \\
& Typographical errors     & \ja{つたい}     & \ja{つらい,辛い} & tough \\
\bottomrule
\end{tabular}
\caption{Word categories extended from \citet{higashiyama-etal-2021-user}. New categories are marked with ``*.'' ``[]'' indicates the context. ${}^\dagger$``\ja{[行く] にゃ}'' is a kitten-style sentence ending and might be expressed as ``[I go,] meow'' or ``[Goin']nya.'' ${}^\ddagger$``\ja{おはこんばんにちは}'' is a coined blend of ``good morning,'' ``good afternoon,'' and ``good evening,'' and might be expressed as ``Good morn-noon-evening.''} \label{tab:cate}
\end{table*}

\subsection{Word Category Definition} \label{sec:cate}
We extended the UGT-specific Japanese word categories defined by \citet{higashiyama-etal-2021-user} and assigned each word in the annotation sentences to every category that it matches. As shown in Table~\ref{tab:cate}, the categories are divided into vocabulary types and variant-form types, with the latter applied non-standard word forms.

\subsection{Standard/Non-Sandard From Definition} \label{sec:sfrom_def}
As stated in \citet{higashiyama-etal-2021-user},
``there are no trivial criteria to determine which
variant forms of a word are standard forms'' (and non-standard forms) ``because most Japanese words can be written in multiple ways.''
Thus, we followed their definition on standard and non-standard forms.
In brief, the definitions can be summarized as follows: standard forms are those variants whose relative frequencies in the reference corpus exceed a set threshold, while non-standard forms are identified per variant category based on falling below category-specific frequency thresholds or exhibiting distinctive orthographic features.
Example non-standard forms for each category is shown in Table~\ref{tab:cate}.
For more detailed definitions, see \S4.2 of their paper.


\begin{table*}[h!]
\centering
\footnotesize
\begin{tabular}{l|rrrr|ccccccc}
\toprule
Data & \multicolumn{4}{c|}{Number} & \multicolumn{7}{c}{F${}_1$ agreement score} \\
& Sent & Word & Cate & VFCate & Word & POS & Lem & Cate & Cate${}^b$ & VFCate & VFCate${}^b$\!\! \\
\midrule
01 \texttt{BJ-OC} & 20 & 327/325 & 30/30 & 19/18 & 99.08 & 98.77 & 98.16 & 83.33 & 83.33 & 86.49 & 86.49 \\
02 \texttt{BJ-OY} & 20 & 359/358 & 52/48 & 17/15 & 99.58 & 96.79 & 95.40 & 86.00 & 94.00 & 75.00 & 81.25 \\
03 \texttt{RC-BLG} & 32 & 1424/1424 & 133/124 & 45/43 & 99.86 & 99.02 & 98.81 & 81.71 & 91.83 & 63.64 & 84.09 \\
04 \texttt{RC-REV} & 41 & 2802/2802 & 129/137 & 22/22 & 100.0 & 99.04 & 98.75 & 96.24 & 96.24 & 100.0 & 100.0 \\
05 \texttt{RK-ICB} & 20 & 600/601 & 27/27 & 15/14 & 99.75 & 98.58 & 98.58 & 85.19 & 96.30 & 82.76 & 89.66 \\
06 \texttt{RK-TRV} & 25 & 748/748 & 23/29 & 11/14 & 100.0 & 99.20 & 98.93 & 84.62 & 84.62 & 80.00 & 80.00 \\
07 \texttt{RK-RCP} & 22 & 442/442 & 43/39 & 26/20 & 99.32 & 98.64 & 97.96 & 87.80 & 92.68 & 82.61 & 82.61 \\
08 \texttt{AM} & 20 & 675/675 & 85/86 & 32/34 & 100.0 & 99.56 & 95.56 & 91.23 & 99.42 & 90.91 & 96.97 \\
09 \texttt{NC-VID} & 20 & 353/358 & 55/50 & 20/10 & 97.33 & 95.08 & 93.67 & 70.48 & 83.81 & 46.67 & 53.33 \\
10 \texttt{NC-PED} & 20 & 443/446 & 63/63 & 24/24 & 98.31 & 97.19 & 94.26 & 77.78 & 85.71 & 62.50 & 66.67 \\
\midrule
Total & 240 & 8173/8179 & 640/633 & 231/214 & 99.66 & 98.65 & 97.82 & 85.78 & 92.22 & 76.85 & 83.60 \\
\bottomrule
\end{tabular}
\caption{Statistics on inter-annotator agreement. The ``Number'' columns show counts of annotated sentences (Sent), words, word categories (Cate), variant-form type categories (VFCate) for each annotator (formatted as ``value1/value2''). The ``F${}_1$ agreement score'' columns report F${}_1$ scores for word segmentation (Seg), parts-of-speech (POS), lemmas (Lem), Cate, and VFCate. Binary agreement on the presence or absence of a category assignment is indicated by ``$b$.''}
\label{tab:iaa}
\end{table*}

\subsection{Standard Form Annotation} \label{sec:align}
The annotation process first identifies word boundaries and then assigns each word to (a category and) a standard form ID, as in the example in Table~\ref{tab:super_tired}; for simplicity, the example displays the standard-form string instead of the ID.
This approach makes it possible to obtain explicit mappings between non-standard words and their standard forms. 

However, when only sentence-level normalization is provided, aligning standard and non-standard forms is not straightforward. In our example, the contiguous span mapping from ``\ja{スッゴクツカレタ～}'' to ``\ja{すごく疲れた}'' (``super tired'') is easily identified, but determining precise word-level alignments remains challenging.

\begin{table}[h]
\small
\centering
\begin{tabular}{ll}
\toprule
Original text & \ja{今日はスッゴクツカレタ～} \\
Segmented text & \ja{今日\,$|$\,は\,$|$\,スッゴク\,$|$\,ツカレタ～}\\
Standard forms &\ja{~~~--~~~\,$|$\,~--~\,$|$\,すごく\,$|$\,疲れた} \\
\bottomrule
\end{tabular}
\caption{An example sentence and its annotation information. The original text means ``Super tired today.''} \label{tab:super_tired}
\end{table}

\subsection{Inter-Annotator Agreement} \label{sec:iaa}
Table~\ref{tab:iaa} shows the detailed statistics on inter-annotator agreement.
Sentences in domains 01--07 were annotated by Annotators B and C, and those in domains 08--10 were by B and D.
Notably, after we fixed annotation disagreements for these sentences through discussions, the sentences were integrated into our final dataset.

\section{Discussion on the Task Definition and Alternative Approaches} \label{sec:task_discussion}
As discussed in \S\ref{sec:task_def}, we adopted a boundary-aware LN task with explicit span identification, which enables direct span-level evaluation.
A simpler text-to-text conversion task can instead be evaluated with sentence-level metrics. However, such metrics (e.g., chrF) allow only relative comparisons: sentences with few normalization targets may achieve high scores even for a trivial “leave-as-is” baseline, indicating that these scores do not reflect absolute output quality.

Moreover, tokenizers used in Japanese encoder-only models may alter characters---for example, splitting a single-character ellipsis ``\ja{…}'' (\texttt{U+2026}) into three consecutive period characters ``.'' (\texttt{U+002E})---causing sentence-level scores to fluctuate even when those characters lie outside normalization spans. Hence, selecting a fair ground-truth sentence—whether the original or a tokenizer-converted sentence—across models is non-trivial.
These considerations impose constraints on character-by-character transduction systems.

One might then ask whether the use of an off-the-shelf word segmenter removes the need to introduce a word boundary prediction step in the encoder-based infilling approach, while still maintaining the feasibility of span-level evaluation. Possible alternatives using an external word segmenter include pre-normalization segmentation and post-normalization segmentation.

On the one hand, the former approach segments the original input text with a segmenter and then normalizes, for example, at the WordPiece level. This preserves word-boundary positions through prefix symbols (e.g., ``ye'' and ``\#\#a'' for ``yea''), similar to \citeauthor{muller-etal-2019-enhancing}’s method for English LN. However, off-the-shelf word segmenters trained on canonical text often produce segmentation errors on informal words, which can directly propagate to the normalization results. 

On the other hand, the latter approach segments the normalizer’s outputs with a segmenter, which generally avoids salient errors when the outputs are accurate. Nevertheless, span-level evaluation under this setting requires non-trivial alignment between ad-hoc word boundaries in the output and those in the original text (see an example in Appendix~\ref{sec:align}).

\section{Encoder-based Approach Variants} \label{sec:enc_variants}
In addition to the \textsc{Full-Seg} approach, we introduce two variant approaches based on the encoder-base detect-and-infill method: \textsc{Part-Seg} and \textsc{Full-Seg-POS}. Experimental results on these variants are reported in \ref{sec:anal_enc}.

\paragraph{Full/Partial Word Segmentation}
In the boundary prediction subtask, we employ full or partial word segmentation (\textsc{Full-Seg} and \textsc{Part-Seg}), depending on whether the training sentences are annotated with full word boundaries or only with non-standard word boundaries.
Unlike the former case described in \S\ref{sec:methods}, in the latter case boundary tags for tokens within standard words should be assigned \verb|O|. For example, the tag sequence should be $[\verb|B|, \verb|I|, \verb|I|, \verb|E|, \verb|O|, \verb|O|, \verb|O|]$ for the input text in Figure~\ref{fig:encoder}.

\paragraph{Word Feature Prediction}
In the detection step, we can optionally employ multi-task learning for word feature prediction.
Specifically, we adopt part-of-speech (POS) tag prediction for each token using an additional linear layer if POS annotation is available.
We refer to this approach as \textsc{Full-Seg-POS}.

\section{Detailed Experimental Settings}
\subsection{Definition of Evaluation Metrics} \label{sec:def_metrics}
Assume each input sentence $\bm{x}$ has a sequence of gold normalization instances $\mathcal{N}_{\bm{x}}=\{(b_i,e_i,S_i)\}_i$, where each instance $(b_i,e_i,S_i)$ consists of a span $(b_i,e_i)$ and a set $S_i=\{s_{i,k}\}_{k=1}^{K_i}$ of one or more standard forms for the corresponding non-standard word.
A system is required to output a set of predicted normalization instances $\hat{\mathcal{N}}_{\bm{x}}=\{(\hat{b}_j,\hat{e}_j,\hat{s}_j)\}_j$, where each $\hat{s}_j$ is a single predicted standard form.

We count a predicted instance as a TP if its span $(\hat{b}_j,\hat{e}_j)$ matches the span $(b_i, e_i)$ of a gold instance and its predicted form $\hat{s}_j$ belongs to the corresponding gold set $S_i$ over all test sentences.
Precision $P$, recall $R$, and the $F_{0.5}$ score over the test set are then defined as follows:
\begin{eqnarray*}
P &=& \frac{\mathrm{TP}}{\mathrm{TP}+\mathrm{FP}},\\
R &=& \frac{\mathrm{TP}}{\mathrm{TP}+\mathrm{FN}},\\
F_{0.5} &=& (1+0.5^2)\frac{PR}{0.5^2 P + R}.
\end{eqnarray*}


\subsection{Pretrained Models} \label{sec:pretrained_models}
We list all pre-trained models used in our experiments in Table~\ref{tab:backbone}.
We selected these models based on their strong performance on general benchmarks; however, for Thai encoder-only and encoder-decoder models, few alternative candidates were available.

\begin{table}[t]
\centering
\footnotesize
\begin{tabular}{ll}
\toprule
Hyperparameter & Value \\
\midrule
Training epochs & ja: 30; th: 20 \\
Batch size & ja: 16; th: 32 \\
Learning rate & 3e-5 \\
Learning rate scheduler & linear \\
Warmup steps & 1 training epoch \\
Gradient norm clipping threshold & 1.0 \\
Optimizer & AdamW \\
\midrule
Training epochs & ja: 30; th: 15 \\
Batch size & ja: 32; th: \{4, 16\} \\
Learning rate & 2e-4 \\
Learning rate scheduler & constant \\
Warmup steps & 1 training epoch \\
Gradient norm clipping threshold & 1.0 \\
Optimizer & AdamW \\
Beam width for inference & 2 \\
\midrule
Training epochs & 10 \\
Batch size & 8 \\
Learning rate & 2e-4 \\
Learning rate scheduler & cosine \\
Warmup ratio & 0.03 \\
Weight decay & 0.001 \\
Gradient norm clipping threshold & 0.3 \\
Optimizer & paged\_adamw\_32bit \\
bf16 & True \\
LoRA rank & 8 \\
LoRA alpha & 16 \\
LoRA dropout & 0.05 \\
LoRA target modules & all linear layers \\
Quantization bit & \{none, 4bit\} \\
Beam width for inference & 1 \\
\bottomrule
\end{tabular}
\caption{Hyperparameter settings for encoder-only (top), encoder-decoder (middle), and decoder-only models (bottom). Batch sizes of 4 and 16 were used for the Thai mT5-large model and the other Thai encoder-decoder models. For models with 7B parameters or more, 4-bit quantization were applied during fine-tuning.}
\label{tab:para_dec}
\end{table}

\subsection{Model Hyperparameters} \label{sec:hypara}
The hyperparameter values used in the experiments are listed in Table \ref{tab:para_dec}.

We conducted hyperparameter search for some important parameters within our computational budgets.
Based on F${}_{0.5}$ score for the JMLN dev set, we chose the best value of learning rate from the search space of \{1e-5, 2e-5, 3e-5, 4e-5, 5e-5\} for encoder-based models using RoBERTa-large, chose the best value of learning rate from \{1e-4, 2e-4, 3e-4, 4e-4, 5e-4\} and that of beam search width during inference from \{1, 2, 3\} for encoder-decoder and decoder-based models using T5-large and Sarashina2-7B, respectively.
We also chose the best value of LoRA rank from \{4, 8, 16\} for decoder-only models using Sarashina2-7B.

\subsection{Prompts for Decoder-only Models} \label{sec:prompt}

We used the instruction prompts in Table~\ref{tab:prompt} as \texttt{\{inst\}} in the full prompt text: ``Instruction:\,{\textbackslash}n\texttt{\{inst\}}{\textbackslash}n{\textbackslash}nInput:\,{\textbackslash}n\texttt{\{src\}}{\textbackslash}n{\textbackslash}nOutput:\,{\textbackslash}n''.
These are common to both Japanese and Thai models, and the \texttt{\{lang\}} placeholder is specified by either language name.

\begin{figure}[h]
\centering
\footnotesize
\begin{tabular}{p{7.25cm}}
\toprule
Instruction text\\
\midrule
If no informal \texttt{\{lang\}} word forms exist in the input text, output the text as is. Otherwise, identify informal word forms and normalize them into their corresponding standard forms. Provide the full normalized text where the original word forms are replaced with the standard forms.\\
\midrule
If no informal \texttt{\{lang\}} word forms exist in the input text, output the text as is. Otherwise, identify informal word forms and normalize them into their corresponding standard forms. Provide the full normalized text, embedding the original and normalized word forms in the format "\texttt{[[}before\texttt{>\,>}after\texttt{]]}". Ensure that the concatenated string of the text outside the brackets and the "before" parts is identical to the input text.\\
\midrule
If no informal \texttt{\{lang\}} word forms exist in the input text, output exactly "NONE". Otherwise, identify every informal word form and normalize it into its corresponding standard form. For each occurrence, output a record in the format "before\texttt{>\,>}after\texttt{>\,>}count". Here, count is the count of how many times the identical original string has already appeared earlier in the input text. If multiple informal forms are found, output each record in the order they occur and separate them with "\texttt{||}".\\
\bottomrule
\end{tabular}
\caption{Instruction prompts for decoder-only models with \textsc{Plain} (top), \textsc{Struct} (middle), and \textsc{Span} approaches (bottom).}
\label{tab:prompt}
\end{figure}

\subsection{Computational Budget for Fine-tuning}
In our experiments, we used NVIDIA V100 GPUs with 32GiB memory, A100 GPUs with 80GiB memory, and H200 GPUs with 140GiB memory.
In total, the models were fine-tuned for 42 GPU hours on V100s, 2700 GPU hours on A100s, and 141 GPU hours on H200s. Encoder-only and encoder-decoder models were fine-tuned on a single GPU, whereas decoder-only models were fine-tuned using eight GPUs.

\subsection{Throughput Calculation Setting} \label{sec:setting_cost}
In the experiments reported in \S\ref{sec:res_cost}, we used the following settings.
For each model checkpoint (from one of two runs), we conducted a single warm-up inference followed by three inference passes over all 3,786 JMLN test sentences (a total of 11.1k characters), which were sorted by increasing token count according to its tokenizer.
These evaluations were run at multiple batch sizes; we selected the batch size that yielded the highest throughput (shown in Table~\ref{tab:cost_batch}) and reported the mean throughput of the three runs. 
Models were cast to \texttt{float16} on the V100 and to \texttt{bfloat16} on the H100.
All inference was performed using the Hugging Face Transformers~\cite{wolf-etal-2020-transformers} library.

\begin{table}[h]
\centering
\small
\begin{tabular}{l|rr|rr}
\toprule
Model & V100 & H200  & V100 & H200 \\
\midrule
& \multicolumn{2}{c|}{\textsc{Full-Seg}} & \\
\midrule
BERT-large    & 64 & 128 & -- & -- \\
RoBERTa-large & 64 & 128 & -- & -- \\
DeBERTa-large & 64 & 128 & -- & -- \\
\midrule
& \multicolumn{2}{c|}{\textsc{Struct}} & \multicolumn{2}{c}{\textsc{Span}} \\
\midrule
T5-large        & 256 & 256 & 256 & 256 \\
Sarashina2.2-1B & 128 & 256 & 256 & 256 \\
Sarashina2.2-3B & 128 & 128 & 64  & 128 \\
Sarashina2-7B   & 64  & 128 & 32  & 128 \\
\bottomrule
\end{tabular}
\caption{Batch size yielding the best throughput for each method.}
\label{tab:cost_batch}
\end{table}

\subsection{Word Coverage Indicators} \label{sec:surf_ind}
For analyses in experiments reported in \S\ref{sec:res_domain} and \S\ref{sec:error_anal}, we introduced three word coverage-based indicators.
Below, we provide further explanations on them.

The Surf-Outside-Train rate is defined as ``the proportion of non-standard surface tokens in the test set that are not found among the training set’s non-standard surfaces.''
Assume that the non-standard surface tokens appearing in the entire training set are [``u'', ``u'', ``r'', ``thx''], while those appearing in the entire test set are [``u'', ``r'', ``cuz'']. In this case, the rate is $1/3=0.33$.

The Surf-In-Train rate is defined as ``the proportion of original surface forms for examples that matched any non-standard forms in the training set.'' Assume the same training set above and non-standard surface examples of interest are [``u'', ``r'', ``cuz''], the rate is $2/3=0.67$

\begin{table*}[t]
\centering
\footnotesize
\begin{tabular}{ll|cc|cc|cc}
\toprule
Data IDs & Register & \multicolumn{2}{c|}{DeBERTa-L} & \multicolumn{2}{c|}{T5-L} & \multicolumn{2}{c}{Sarashina2.2-3B} \\
&& P & R & P & R & P & R \\
\midrule
01        & Q\&A site    & 0.654 & 0.416 & 0.572 & 0.369 & \textbf{0.717} & \textbf{0.578} \\
02--03    & Blog         & 0.750 & 0.595 & 0.761 & 0.591 & \textbf{0.802} & \textbf{0.685} \\
04--06,08 & Reviews      & \textbf{0.837} & 0.630 & 0.769 & 0.568 & 0.808 & \textbf{0.687} \\
07        & Recipes      & 0.815 & 0.684 & 0.817 & 0.663 & \textbf{0.834} & \textbf{0.734} \\
09        & Video desc.  & 0.627 & 0.538 & 0.666 & 0.492 & \textbf{0.730} & \textbf{0.629} \\
10        & Forum        & \textbf{0.850} & 0.747 & 0.783 & 0.677 & 0.808 & \textbf{0.772} \\
11        & Social media & 0.613 & 0.511 & 0.600 & 0.464 & \textbf{0.763} & \textbf{0.677} \\
12        & Wiki hist.   & \textbf{0.528} & 0.140 & 0.197 & 0.095 & 0.509 & \textbf{0.280} \\
13--14    & Conv. trans. & \textbf{0.816} & 0.662 & 0.699 & 0.537 & 0.812 & \textbf{0.710} \\
\bottomrule
\end{tabular}
\caption{JMLN test results of representative three models for each aggregated domain.} \label{tab:res_ja_domain_agg}
\end{table*}

The Norm-In-Lex rate is defined as ``the proportion of predicted normalized forms appearing in the UniDic lexicon.''
Assume the set of normalized strings produced by the model is \{``you'', ``are'', ``becaus''\}, and an external lexicon contains only ``you'' and ``are'' from that set; the rate is $2/3=0.67$.

\section{Detailed Experimental Results} \label{sec:detailed_exp}
\subsection{Accuracy Across Model Series} \label{sec:model_series_figure}
\begin{figure}[h]
\centering
\includegraphics[width=\linewidth]{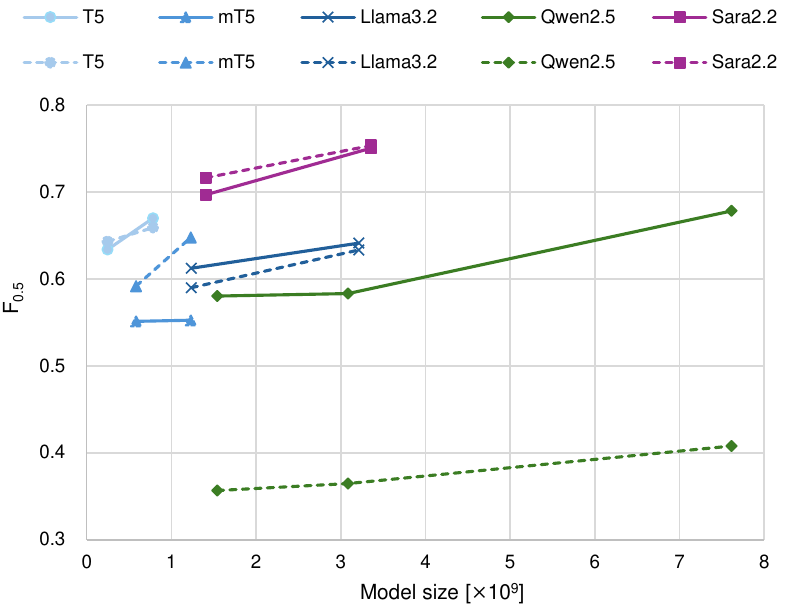}
\caption{Plot of F${}_{0.5}$ scores in Table\ref{tab:res_ja} for each model series---T5, mT5, Llama-3.2, Qwen2.5, and Sarashina2.2. The scores for the \textsc{Struct} and \textsc{Span} approaches are shown with solid and dotted lines, respectively.}
\label{fig:model_series}
\end{figure}

For the experiment reported in Table\ref{tab:res_ja} (\S\ref{sec:res_acc_ja}),
we extracted results for different model size within each model series and plotted them in Figure~\ref{fig:model_series}.
Performance improved with increasing model size across model series.

\subsection{Results Across Grouped Domains} \label{sec:res_domain_agg}
Table~\ref{tab:res_ja_domain_agg} shows the performance of three selected models as in \S\ref{sec:res_domain}---DeBERTa-large, T5-large (\textsc{Span}), and Sarashina-2.2-3B (\textsc{Span})---on the grouped domain test set of JMLN, in which domains of a similar type from different data sources (e.g., all blog sites) are combined into a single domain group.
The results are consistent with those on the single-domain test sets.
Sarashina-2.2-3B achieved the highest recall across all grouped domains and also obtains the highest precision in 5 out of 9 groups, while DeBERTa-large achieved the highest precision in the remaining groups.

\subsection{Comparison of Encoder Approaches} \label{sec:anal_enc}
As a preliminary experiment, we compared variants of encoder-based methods---\textsc{Part-Seg}, \textsc{Full-Seg}, and \textsc{Full-Seg-POS}, introduced in Appendix~\ref{sec:enc_variants}---on the JMLN dev set.
We also report the F1 scores for word segmentation, POS tagging, and length prediction (Len${}^p$: tokens in positive non-standard words; Len${}^n$: other tokens), as well as additional normalization metrics: sentence-level exact match accuracy (S-Acc${}^p$: accuracy for sentences containing at least one positive non-standard word; S-Acc${}^n$: accuracy for other sentences) and the chrF score~\cite{popovic-2015-chrF} implemented in sacreBLEU~\cite{post-2018-call}.\footnote{We calculated chrF scores with the default options (signature: \texttt{nrefs:1|case:mixed|eff:yes|nc:6|nw:0|}
\texttt{space:no|version:2.5.1}).}
Table~\ref{tab:res_ja_enc_detail} shows the results.

The encoder-only models (3 backbone models$\times$2 model size) with the \textsc{Full-Seg} approach obtained +0.01--0.05 F${}_{0.5}$ gains from the \textsc{Part-Seg} counterpart, indicating the importance of learning full word segmentation tasks.
Adding POS tagging task showed no clear improvements in many cases.

\begin{table*}[t]
\centering
\small
\begin{tabular}{ll|ccc|ccc|cc}
\toprule
Backbone & Approach & \multicolumn{3}{c|}{Det} & \multicolumn{3}{c|}{Norm} & \multicolumn{2}{c}{Det$-$Norm}\\
&& P & R & F${}_{0.5}$ & P & R & F${}_{0.5}$ & P & R \\
\midrule
RoBERTa-base & \textsc{Full-Seg} & \textbf{0.804} & 0.596 & \textbf{0.751} & \textbf{0.713} & 0.529 & \textbf{0.666} & 0.091 & 0.067 \\
\midrule
T5-base & \textsc{Span} & 0.715 & 0.684 & 0.708 & 0.645 & 0.618 & 0.640 & 0.069 & 0.066 \\
mT5-large & \textsc{Span} & 0.725 & 0.670 & 0.713 & 0.660 & 0.609 & 0.649 & 0.066 & 0.061 \\
\midrule
Llama3.2-Typhoon2-3b & \textsc{Span} & 0.717 & 0.731 & 0.720 & 0.656 & 0.668 & 0.658 & 0.062 & 0.063 \\
Llama3.2-Typhoon2-8b & \textsc{Span} & 0.722 & \textbf{0.741} & 0.726 & 0.661 & \textbf{0.678} & 0.664 & 0.061 & 0.063 \\
\bottomrule
\end{tabular}
\caption{VISTEC test results of selected Thai models for detection (det) and normalization (norm) tasks. The best score among compared systems is highlighted in \textbf{bold}.
}
\label{tab:res_th_det}
\end{table*}

\subsection{Comparison of Generative Approaches} \label{sec:anal_plain}
As a preliminary experiment, we compared the three generative approaches---\textsc{Plain}, \textsc{Struct}, and \textsc{Span}--- using T5, Sarashina2/2.2, and Swallow models on the JMLN dev set.
The \textsc{Struct} and \textsc{Span} approaches outperformed or matched the \textsc{Plain} approach for T5 and Sarashina2/2.2 models (e.g., \mbox{-0.009} to +0.034 S-Acc${}^p$ points), but underperformed for Swallow (e.g., -0.076 S-Acc${}^p$ points).

Notably, we also trained a character-level Transformer based on the T5 architecture (the same size as T5-efficient-tiny\footnote{\url{https://huggingface.co/google/t5-efficient-tiny}}) from scratch with full JMLN training set, but it failed to reach adequate performance ($\approx$14 chrF on the dev set). 

\subsection{Informal Span Detection Accuracy for Thai} \label{sec:det_thai}
Table~\ref{tab:res_th_det} shows the performance of the informal word span detection task for selected Thai models on the VISTEC test set.
Consistent with the Japanese results in \S\ref{sec:exp_detect}, the encoder-only model exhibited larger precision/recall differences between the two tasks, while the recall differences remained similar across the three model types.
Overall, the high-precision encoder-only model and the high-recall encoder-decoder and decoder-only models yielded in F${}_{0.5}$ scores at a comparable level.

\subsection{Results Across Category}
Table~\ref{tab:res_ja_cate} shows the JMLN test performance (recall) of the three models evaluated in \S\ref{sec:res_domain} for each variant-form type category.
Consistent with the domain-specific evaluation results in \S\ref{sec:res_domain}, Sarashina2.2-3B achieved the highest recall across all categories.
All three models showed the same recall order across the four categories, and the consistently lowest recall of the fourth category, i.e., typos, once again highlights the difficulty of correcting them.

\begin{table}[h]
\centering
\footnotesize
\begin{tabular}{lc|ccc}
\toprule
Category & \# & DeBERTa & T5 & Sarashina \\
\midrule
Char type var.   & 534 & 0.395 & 0.375 & \textbf{0.563} \\
Alter rep.       & 271 & 0.550 & 0.500 & \textbf{0.703} \\
Sound change     & 794 & 0.796 & 0.739 & \textbf{0.812} \\
Typos            & 178 & 0.087 & 0.076 & \textbf{0.219} \\
\bottomrule
\end{tabular}
\caption{JMLN test recall of representative models (DeBERTa-large, T5-large, and Sarashina2.2-3B) for each variant-form type category.} \label{tab:res_ja_cate}
\end{table}

\subsection{Prediction Examples} \label{sec:error_examples}
Table~\ref{tab:error_examples} shows model outputs on the JMLN dev set for the three models evaluated in \S\ref{sec:res_domain}; in addition to error type classification (TP, FP, and FN) based on the gold standard evaluation, the first author also assessed the validity of each output ($\checkmark$: valid, $\triangle$: questionable, \xmark: invalid).

In example (a), all three methods produced semantically valid normalized strings at the phrase level; however, only DeBERTa was classified as a TP, as it correctly matched the gold span.
Similarly, in examples (b) and (c), Sarashina produced plausible normalized strings, but due to mismatches with the gold annotation, the outputs were classified as FPs.
In example (d), Sarashina normalized the input to a synonym (i.e., ``darling'' to ``husband''); while the result was semantically appropriate, it falls outside the scope of the task, which requires normalization to variant surface forms of the same word.
In examples (e) and (f), Sarashina produced normalized strings that were semantically unrelated to the original input.
DeBERTa generated invalid, non-word outputs in examples (b) and (d), while T5 did so in example (c).
Example (g) shows that all three models produced different but erroneous outputs, suggesting that normalization becomes particularly challenging when multiple non-standard words appear in sequence, making boundary and word identification more difficult.

\begin{table*}[t]
\centering
\small
\begin{tabular}{lc|cccc|cccccc}
\toprule
Backbone & Approach & Seg & POS & Len${}^p$ & Len${}^n$ & \multicolumn{6}{c}{Norm} \\
&& F${}_1$ & F${}_1$ & F${}_{1}$ & F${}_{1}$ & P & R & F${}_{0.5}$ & S-Acc${}^p$ & S-Acc${}^n$ & chrF \\
\midrule
\midrule
\multicolumn{2}{l|}{Leave-As-Is} & -- & -- & -- & -- & -- & 0 & -- & 0 & 1 & 95.29 \\
\midrule
\multirow{3}{*}{BERT-B}
& \textsc{P-Seg} & --    & --    & 0.644 & --    & 0.676 & 0.510 & 0.635 & 0.487 & 0.982 & 96.69 \\
& \textsc{F-Seg}  & 0.980 & --    & \underline{0.656} & 0.978 & \underline{0.721} & \underline{0.546} & \underline{0.678} & 0.518 & 0.979 & 96.82 \\
& \textsc{F-Seg-POS} & 0.978 & 0.963 & 0.636 & 0.976 & 0.649 & 0.532 & 0.621 & 0.508 & 0.970 & 96.56 \\
\midrule
\multirow{3}{*}{RoBERTa-B}
& \textsc{P-Seg} & --    & --    & 0.630 & --    & 0.662 & 0.515 & 0.626 & 0.488 & 0.976 & 97.04 \\
& \textsc{F-Seg}  & 0.982 & --    & 0.669 & 0.980 & \underline{0.726} & 0.537 & \underline{0.678} & 0.502 & 0.981 & 97.22 \\
& \textsc{F-Seg-POS} & 0.982 & 0.971 & \underline{0.677} & 0.981 & 0.713 & \underline{0.566} & \underline{0.678} & 0.525 & 0.973 & 97.27 \\
\midrule
\multirow{3}{*}{DeBERTa-B}
& \textsc{P-Seg} & --    & --    & 0.637 & --    & 0.678 & 0.516 & 0.638 & 0.489 & 0.977 & 97.12 \\
& \textsc{F-Seg}  & 0.982 & --    & {0.656} & 0.980 & \underline{0.745} & 0.516 & \underline{0.685} & 0.477 & 0.989 & 97.16 \\
& \textsc{F-Seg-POS} & 0.983 & 0.972 & \underline{0.680} & 0.982 & 0.716 & \underline{0.566} & 0.680 & 0.528 & 0.974 & 97.26 \\
\midrule
\multirow{3}{*}{BERT-L}
& \textsc{P-Seg} & --    & --    & 0.682 & --    & 0.717 & 0.570 & 0.682 & 0.542 & 0.979 & 96.92 \\
& \textsc{F-Seg}  & 0.984 & --    & \underline{0.704} & 0.983 & \underline{0.769} & \underline{0.588} & \underline{0.725} & 0.549 & 0.979 & 97.01 \\
& S\&P & 0.980 & 0.967 & 0.691 & 0.979 & 0.756 & 0.576 & 0.712 & 0.534 & 0.984 & 97.04 \\
\midrule
\multirow{3}{*}{RoBERTa-L}
& \textsc{P-Seg} & --    & --    & 0.671 & --    & 0.706 & 0.557 & 0.670 & 0.515 & 0.980 & 97.27 \\
& \textsc{F-Seg}  & 0.983 & --    & 0.701 & 0.982 & 0.753 & 0.571 & 0.708 & 0.542 & 0.980 & 97.34 \\
& \textsc{F-Seg-POS} & 0.985 & 0.974 & \underline{0.713} & 0.983 & \underline{0.765} & \underline{0.580} & \underline{0.719} & 0.551 & 0.984 & 97.49 \\
\midrule
\multirow{3}{*}{DeBERTa-L}
& \textsc{P-Seg} & --    & --    & 0.685 & --    & 0.763 & 0.556 & 0.710 & 0.523 & 0.985 & 97.34 \\
& \textsc{F-Seg}  & 0.986 & --    & 0.718 & 0.984 & 0.763 & \underline{0.593} & 0.722 & 0.556 & 0.980 & 97.54 \\
& \textsc{F-Seg-POS} & 0.985 & 0.975 & \underline{0.727} & 0.983 & \underline{0.788} & 0.592 & \underline{0.739} & 0.548 & 0.984 & 97.50 \\
\bottomrule
\end{tabular}
\caption{Results of encoder-only models with different approaches (\textsc{P-Seg}: \textsc{Part-Seg}, \textsc{F-Seg}: \textsc{Full-Seg}, and \textsc{F-Seg-POS}: \textsc{Full-Seg-POS}) on the JMLN dev sets. For each model, the best score among its variants is \underline{underlined}.}
\label{tab:res_ja_enc_detail}
\end{table*}

\begin{table*}[h]
\centering
\small
\begin{tabular}{ll|cccccc}
\toprule
Backbone & Approach & P & R & F${}_{0.5}$ & S-Acc${}^p$ & S-Acc${}^n$ & chrF \\
\midrule
-- & Leave-As-Is & -- & 0 & -- & 0 & 1 & 95.29 \\
\midrule
\multirow{3}{*}{T5-base}
& \textsc{Plain} & --    & --    & --    & 0.497 & 0.958 & 96.54 \\
& \textsc{Struct}& \underline{0.727} & 0.491 & \underline{0.663} & 0.465 & \underline{0.983} & 96.94 \\
& \textsc{Span}  & 0.707 & \underline{0.527} & 0.662 & \underline{0.509} & {0.974} & \underline{97.01} \\
\midrule
\multirow{3}{*}{T5-large} 
& \textsc{Plain} & --    & --    & --    & 0.535 & 0.950 & 96.44 \\
& \textsc{Struct}& \underline{0.769} & 0.558 & \underline{0.714} & 0.531 & \underline{0.984} & \underline{97.31} \\
& \textsc{Span}  & 0.751 & \underline{0.568} & 0.705 & \underline{0.566} & 0.975 & 97.28 \\
\midrule
\multirow{3}{*}{Sarashina2.2-1B} 
& \textsc{Plain} & -- & -- & -- & \underline{0.602} & \underline{0.985} & \underline{97.92} \\
& \textsc{Struct}& 0.750 & \underline{0.635} & 0.724 & \underline{0.602} & 0.972 & 97.72 \\
& \textsc{Span}  & \underline{0.770} & 0.609 & \underline{0.730} & 0.593 & 0.975 & 97.59 \\
\midrule
\multirow{3}{*}{Sarashina2.2-3B}
& \textsc{Plain} & -- & -- & -- & \underline{0.656} & \underline{0.980} & \underline{98.17} \\
& \textsc{Struct}& \underline{0.792} & \underline{0.682} & 0.766 & 0.651 & 0.972 & 97.94 \\
& \textsc{Span}  & \underline{0.792} & 0.677 & \underline{0.767} & 0.653 & 0.979 & 98.11 \\
\midrule
\multirow{3}{*}{Sarashina2-7B}
& \textsc{Plain} & -- & -- & -- & 0.646 & \underline{0.982} & \underline{98.08} \\
& \textsc{Struct}& 0.766 & \underline{0.670} & \underline{0.745} & \underline{0.658} & 0.971 & 97.99 \\
& \textsc{Span}  & \underline{0.767} & 0.662 & 0.743 & 0.637 & 0.965 & 97.74 \\
\midrule
\multirow{3}{*}{Llama-3.1-Swallow-8B}
& \textsc{Plain} & -- & -- & -- & \underline{0.638} & \underline{0.983} & \underline{98.03} \\
& \textsc{Struct}& \underline{0.771} & \underline{0.609} & \underline{0.732} & 0.584 & 0.980 & 97.68 \\
& \textsc{Span}  & 0.755 & 0.602 & 0.718 & 0.562 & 0.974 & 97.50 \\
\bottomrule
\end{tabular}
\caption{Normalization results of encoder-decoder and decoder-only models on the JMLN dev sets. For each model, the best score among its variants is \underline{underlined}.}
\label{tab:res_ja_gen_detail}
\end{table*}

\begin{table*}[h]
\centering
\footnotesize
\begin{tabular}{cllccc}
\toprule
\multirow{6}{*}{(a)}
& \multicolumn{5}{l}{\texttt{RK-ICB}: $(\cdots)$\,\ja{効果はわから\underline{ない〜〜}${}_{\mathsf{Pos:Unk}}$。}} \\
& \multicolumn{5}{l}{\qquad\qquad (\underline{Not} sure about the effect.)} \\
\cmidrule{2-6}
& Gold       & \ja{ない〜〜}$\rightarrow$\{\ja{ない}\} & (not) & \\
& DeBERTa-L  & \ja{ない〜〜}$\rightarrow$\ja{\underline{ない}${}_{\mathsf{InL}}$} & (not) & TP & $\checkmark$ \\
& T5-L       & \ja{わからない〜〜}$\rightarrow$\ja{\underline{わからない}${}_{\mathsf{OOL}}$} & (not sure) & FP\&FN & $\checkmark$ \\
& Sarashina2.2-3B & \ja{わからない〜〜}$\rightarrow$\ja{\underline{わからない}${}_{\mathsf{OOL}}$} & (not sure) & FP\&FN & $\checkmark$ \\
\midrule
\multirow{6}{*}{(b)}
& \multicolumn{5}{l}{\texttt{BJ-OC}: \ja{かいわれ大根って、$(\cdots)$\,水で洗ったらそのまま\underline{だべれる}${}_{\mathsf{Pos:Unk}}$の?}} \\
& \multicolumn{5}{l}{\qquad\qquad (\underline{Can} you just was daikon radish sprouts with water and \underline{eat} it as is?)} \\
\cmidrule{2-6}
& Gold       & \ja{だべれる}$\rightarrow$\{\ja{食べれる}\} & (can eat) & \\
& DeBERTa-L  & \ja{だべれる}$\rightarrow$\ja{\underline{食れる}${}_{\mathsf{OOL}}$} & ($\varnothing$) & FP\&FN & \xmark \\
& T5-L       & -- & & FN & \xmark \\
& Sarashina2.2-3B & \ja{だべれる}$\rightarrow$\ja{\underline{食べられる}${}_{\mathsf{OOL}}$} & (can eat) & FP\&FN & $\checkmark$ \\
\midrule
\multirow{6}{*}{(c)}
& \multicolumn{5}{l}{\texttt{BJ-OC}: \ja{$(\cdots)$\,近鉄の「まわりゃんせ」が\underline{おトク}${}_{\mathsf{Neg:Unk}}$じゃないでしょうか。}} \\
& \multicolumn{5}{l}{\qquad\qquad (Isn't Kintetsu Railway's \textit``Mawaryanse'' pass a \underline{good deal}?)} \\
\cmidrule{2-6}
& Gold       & -- & & \\
& DeBERTa-L  & -- & & -- &  --\\
& T5-L       & \ja{トク}$\rightarrow$\ja{\underline{とこう}${}_{\mathsf{InL}}$} & ($\varnothing$) & FP & \xmark \\
& Sarashina2.2-3B & \ja{おトク}$\rightarrow$\ja{\underline{お得}${}_{\mathsf{OOL}}$} & (good deal) & FP & $\checkmark$ \\
\midrule
\multirow{6}{*}{(d)}
& \multicolumn{5}{l}{\texttt{BJ-OY}: \ja{$(\cdots)$\,\underline{だぁりん}${}_{\mathsf{Pos:Unk}}$がおごってくれました。}} \\
& \multicolumn{5}{l}{\qquad\qquad (My \underline{darling} treated me.)} \\
\cmidrule{2-6}
& Gold       & \ja{だぁりん}$\rightarrow$\{\ja{ダーリン}\} & (darling) & \\
& DeBERTa-L  & \ja{だぁりん}$\rightarrow$\ja{\underline{だじりん}${}_{\mathsf{OOL}}$} & ($\varnothing$)    & FP\&FN & \xmark \\
& T5-L       & -- && FN & \xmark \\
& Sarashina2.2-3B & \ja{だぁりん}$\rightarrow$\ja{\underline{旦那}${}_{\mathsf{InL}}$} & (husband) & FP\&FN & $\triangle$ \\
\midrule
\multirow{7}{*}{(e)}
& \multicolumn{5}{l}{\texttt{BJ-OY}: \ja{$(\cdots)$\,ウォルトンで\underline{カンツリ}${}_{\mathsf{Neg:Unk}}$と利根川水系でバスをやんべ。}} \\
& \multicolumn{5}{l}{\qquad\qquad (I'm going \underline{kan-tsuri} (=\,fishing at a managed sport) at Walton and bass fishing } \\
& \multicolumn{5}{l}{\qquad\qquad in Tone River system.)} \\
\cmidrule{2-6}
& Gold       & -- & & \\
& DeBERTa-L  & -- & & -- & -- \\
& T5-L       & -- & & -- & -- \\
& Sarashina2.2-3B & \ja{カンツリ}$\rightarrow$\ja{\underline{キャッチ＆リリース}${}_{\mathsf{OOL}}$} & (catch \& release) & FP & \xmark \\
\midrule
\multirow{6}{*}{(f)}
& \multicolumn{5}{l}{\texttt{NU}: \ja{$(\cdots)$\,最高気温25度とか\underline{っしょう}${}_{\mathsf{Pos:Unk}}$?}} \\
& \multicolumn{5}{l}{\qquad\qquad (The high is like 25\ja{℃} or something, \underline{right}?)} \\
\cmidrule{2-6}
& Gold       & \ja{っしょう}$\rightarrow$\{\ja{でしょう}\} & (right?) & \\
& DeBERTa-L  & -- & & FN & \xmark \\
& T5-L       & -- & & FN & \xmark \\
& Sarashina2.2-3B & \ja{かっしょう}$\rightarrow$\ja{\underline{夏}${}_{\mathsf{InL}}$} & (summer) & FP\&FN & \xmark \\
\midrule
\multirow{6}{*}{(g)}
& \multicolumn{5}{l}{\texttt{TW}: \ja{あ〜あ\underline{まぁ〜た}${}_{\mathsf{Pos:Unk}}$初音ミクがトレンド入りしてるよ\,$(\cdots)$}} \\
& \multicolumn{5}{l}{\qquad\qquad (Ugh, Hatsune Miku is trending \underline{again}.)} \\
\cmidrule{2-6}
& Gold       & \ja{まぁ〜た}$\rightarrow$\{\ja{また}\} & (again) & \\
& DeBERTa-L  & \ja{あまぁ〜}$\rightarrow$\ja{\underline{あまっ}${}_{\mathsf{InL}}$} & (extra) & FP\&FN & \xmark \\
& T5-L       & \ja{まぁ〜た}$\rightarrow$\ja{\underline{まあ}${}_{\mathsf{InL}}$} & (well...) & FP\&FN & \xmark \\
& Sarashina2.2-3B & \ja{あまぁ〜た}$\rightarrow$\ja{\underline{あまった}${}_{\mathsf{OOL}}$} & (extra) & FP\&FN & \xmark \\
\bottomrule
\end{tabular}
\caption{Example original text (fragment) and corresponding model outputs. For each row of gold-standard and model output, columns 2--5 indicate: (1) original string$\rightarrow$normalized string, (2) gloss of the normalized string, (3) error type, (4) manual validity judgement by the first author ($\checkmark$: valid, $\triangle$: questionable, \xmark: invalid). Non-standard words in the original text are underlined; if positive instances, they are marked with a subscript $\mathsf{Pos:Unk}$, and if negative, with $\mathsf{Neg:Unk}$ (Unk indicates that the word did not appear in the training set). Predicted normalized strings are marked with subscripts $\mathsf{InL}$ or $\mathsf{OOL}$, indicating whter the form is included or not included in the UniDic lexicon.} \label{tab:error_examples}
\end{table*}

\begin{table*}[t!]
\centering
\footnotesize
\begin{tabular}{llcccrll}
\toprule
ID & Pretrained Model & Lang & \multicolumn{2}{c}{Exp} & Size & Hugging Face ID & License \\
&&& ja & th &&& \\
\midrule
01 & BERT-base    & ja & $\checkmark$ &  & 91M & tohoku-nlp/bert-base-japanese-char-v3 & Apache 2.0\\
02 & BERT-large   & ja & $\checkmark$ & & 310M & tohoku-nlp/bert-large-japanese-char-v2 & Apache 2.0\\
03 & RoBERTa-base & ja & $\checkmark$ & & 100M & ku-nlp/roberta-base-japanese-char-wwm & CC BY-SA 4.0\\
04 & RoBERTa-large& ja & $\checkmark$ & & 320M & ku-nlp/roberta-large-japanese-char-wwm & CC BY-SA 4.0 \\
05 & RoBERTa-base & th & & $\checkmark$ & 88M & KoichiYasuoka/roberta-base-thai-char & Apache 2.0\\
06 & DeBERTa-base & ja & $\checkmark$ & & 100M & ku-nlp/deberta-v2-base-japanese-char-wwm & CC BY-SA 4.0\\
07 & DeBERTa-large& ja & $\checkmark$ & & 330M & ku-nlp/deberta-v2-large-japanese-char-wwm & CC BY-SA 4.0\\
\midrule
08 & T5-base  & ja & $\checkmark$ & & 250M & retrieva-jp/t5-base-long & CC BY-SA 4.0\\
09 & T5-large & ja & $\checkmark$ & & 780M & retrieva-jp/t5-large-long & CC BY-SA 4.0\\
10 & T5-base  & th & & $\checkmark$ & 250M & kobkrit/thai-t5-base & N/A \\
11 & mT5-base & M & $\checkmark$ & $\checkmark$ & 580M & google/mt5-base & Apache 2.0\\
12 & mT5-large& M & $\checkmark$ & $\checkmark$ & 1.2B & google/mt5-large & Apache 2.0\\
\midrule
13 & Llama-3.2-1B    & M & $\checkmark$ & $\checkmark$ & 1.2B & meta-llama/Llama-3.2-1B & Llama 3.2 \\
14 & Llama-3.2-3B    & M & $\checkmark$ & $\checkmark$ & 3.2B & meta-llama/Llama-3.2-3B & Llama 3.2 \\
15 & Llama-3.1-8B    & M & $\checkmark$ & $\checkmark$ & 8.0B & meta-llama/Llama-3.1-8B & Llama 3.1 \\
16 & Swallow-8B & M$\rightarrow$ja & $\checkmark$ & & 8.0B & tokyotech-llm/Llama-3.1-Swallow-8B-v0.2 & Llama 3.1 \\
17 & Qwen2.5-1.5B   & M & $\checkmark$ & $\checkmark$ & 1.5B & Qwen/Qwen2.5-1.5B & Apache 2.0\\
18 & Qwen2.5-3B     & M & $\checkmark$ & $\checkmark$ & 3.1B & Qwen/Qwen2.5-3B & Qwen Research \\
19 & Qwen2.5-7B     & M & $\checkmark$ & $\checkmark$ & 7.6B & Qwen/Qwen2.5-7B & Apache 2.0\\
20 & TinySwallow-1.5B& M$\rightarrow$ja & $\checkmark$ & & 1.5B & SakanaAI/TinySwallow-1.5B & Apache 2.0 \\
21 & Sarashina2.2-1B & ja\&en & $\checkmark$ & & 1.4B & sbintuitions/sarashina2.2-1b & MIT\\
22 & Sarashina2.2-3B & ja\&en & $\checkmark$ & & 3.4B & sbintuitions/sarashina2.2-3b & MIT\\
23 & Sarashina2-7B   & ja\&en & $\checkmark$ & & 7.3B & sbintuitions/sarashina2-7b & MIT\\
24 & Typhoon2-1B     & M$\rightarrow$th & & $\checkmark$ & 1.2B & scb10x/llama3.2-typhoon2-1b & Llama 3.2\\
25 & Typhoon2-3b     & M$\rightarrow$th & & $\checkmark$ & 3.2B & scb10x/llama3.2-typhoon2-3b & Llama 3.2\\
26 & Typhoon2-8B     & M$\rightarrow$th & & $\checkmark$ & 8.0B & scb10x/llama3.1-typhoon2-8b & Llama 3.1\\
27 & ThaiGPT1.5-7B   & M$\rightarrow$th & & $\checkmark$ & 7.6B & openthaigpt/openthaigpt1.5-7b-instruct & Qwen \\
28 & SeaLLMs-v3-1.5B & M$\rightarrow$M & & $\checkmark$ & 1.5B & SeaLLMs/SeaLLMs-v3-1.5B & SeaLLMs \\
29 & SeaLLMs-v3-7B   & M$\rightarrow$M & & $\checkmark$ & 7.6B & SeaLLMs/SeaLLMs-v3-7B & SeaLLMs \\
\bottomrule
\end{tabular}
\caption{Backbone models used in Japanese and Thai experiments (``Exp''=``ja'' and ``th''). The ``Lang'' column indicates the languages that the model mainly trained on (``M'' indicates a multilingual model, and ``M$\rightarrow$*'' indicates a continually pre-trained model from a multilingual model).}
\label{tab:backbone}
\end{table*}

\end{document}